\documentclass[a4paper,fleqn,review]{cas-sc}

\usepackage[utf8]{inputenc}
\usepackage{longtable}
\usepackage{float}
\usepackage{enumitem}
\setlist{nosep}

\usepackage[numbers,sort&compress]{natbib}

\newcommand{\hd}[1]{\begin{tabular}[b]{@{}c@{}}#1\end{tabular}}

\ExplSyntaxOn
\cs_set:Npn \__make_tbl_caption:nn #1#2
{
  \l_tbl_align_tl
  \skip_vertical:N \l_tbl_abovecap_skip
  {\parbox{ \dimexpr(\l_tbl_width_dim)}
    {\rightskip=0pt\sffamily\footnotesize\textbf{\color{scolor}#1}\par#2\par\vskip4pt }}
  \skip_vertical:N \l_tbl_belowcap_skip
}
\cs_set:Npn \__make_fig_caption:nn #1#2
{
  \l_fig_align_tl
  \skip_vertical:N \l_fig_abovecap_skip
  \setbox\cascaptionbox=\hbox{%
     \sffamily\footnotesize\textbf{\color{scolor}#1:}~#2}
  \ifdim\the\wd\cascaptionbox<\dim_use:N \l_fig_width_dim\relax
    \parbox{ \l_fig_width_dim }
      {\unskip\ignorespaces\hfil\sffamily\footnotesize
       \textbf{\color{scolor}#1:}~#2\hfil\par }
  \else
    \parbox{ \l_fig_width_dim }
      {\rightskip=0pt\unskip\ignorespaces\sffamily
       \footnotesize\textbf{\color{scolor}#1:}~#2\par }
  \fi
  \skip_vertical:N \l_fig_belowcap_skip
}
\makeatletter
\RenewDocumentEnvironment { table } { O{} }
{
  \__reset_tbl:
  \bool_gset_false:N \g_tbl_full_bool
  \dim_set:Nn \l_tbl_width_dim { \linewidth }
  \keys_set:nn { cas / tbl } { #1 }
  \csxdef{fps@table}{\l_tbl_pos_tl}
  \csgdef{tblwidth}{\dim_use:N \l_tbl_width_dim}
  \cs_set_eq:NN \@makecaption \__make_tbl_caption:nn
  \@float{table}
  \l_tbl_align_tl
  \sffamily\normalsize
}
{ \end@float }
\RenewDocumentEnvironment { table* } { O{width=\FullWidth} }
{
  \__reset_tbl:
  \bool_gset_true:N \g_tbl_full_bool
  \dim_set:Nn \l_tbl_width_dim { \FullWidth }
  \keys_set:nn { cas / tbl } { #1 }
  \csxdef{fps@table}{\l_tbl_pos_tl}
  \csgdef{tblwidth}{\dim_use:N \l_tbl_width_dim}
  \cs_set_eq:NN \@makecaption \__make_tbl_caption:nn
  \@dblfloat{table}
  \l_tbl_align_tl
  \sffamily\normalsize
}
{ \end@dblfloat }
\makeatother
\bool_gset_true:N \g_stm_nologo_bool
\ExplSyntaxOff

\begin{document}
\let\WriteBookmarks\relax
\def\floatpagepagefraction{1}
\def\textpagefraction{.001}
\shorttitle{AutoEnergyDesign}
\shortauthors{A. Buonanno et~al.}

\title [mode = title, size = \fontsize{14pt}{17pt}\selectfont]{AutoEnergyDesign: An LLM-Driven Agent for Automatic Energy System Design Optimization}

\author[1]{Amedeo Buonanno}[orcid=0000-0003-3494-2648]
\cormark[1]
\ead{amedeo.buonanno@enea.it}
\credit{Conceptualization, Methodology, Software, Investigation, Data curation, Visualization, Writing -- original draft, Writing -- review \& editing}

\author[1]{Salvatore Fabozzi}
\credit{Conceptualization, Investigation, Data curation, Validation, Writing -- original draft, Writing -- review \& editing}

\author[1]{Maria Valenti}
\credit{Conceptualization, Resources, Writing -- original draft, Writing -- review \& editing}

\affiliation[1]{organization={Department of Energy Technologies and Renewable Sources, ENEA},
	      address={P.le Enrico Fermi, 1},
                postcode={80055},
                city={Portici (NA)},
                country={Italy}}

\cortext[cor1]{Corresponding author}

\begin{abstract}
The design optimization of building energy systems represents a complex, high-dimensional search problem involving discrete technology selections, continuous sizing parameters, and intricate economic trade-offs in which every candidate must be evaluated by a year-long simulation. This paper presents \textit{AutoEnergyDesign}, in which a Large Language Model (LLM) agent reasons over past simulation results, proposes configuration changes and learns in context from each outcome, while a git-based mechanism reverts changes that do not improve the design.
The agent couples a reasoning model with explicit memories, a constrained action interface, and an environment-grounded keep/revert feedback signal.
The simulator covers eight technologies and computes the 25-year Net Present Cost (NPC).
On a case study with measured building loads, the agent, instantiated with three Claude models (Haiku~5.5, Sonnet~5.5 and Opus~5.5; ten runs of 50 simulations each), is compared with random search, Tree-structured Parzen Estimator (TPE) Bayesian optimization and a $(1+1)$ evolution strategy that shares the agent's acceptance rule (ten seeds of 300 evaluations). With all three models the agent reaches an NPC of 3.53~MEUR, 1.20\% and 0.23\% below the best results of TPE and of the evolution strategy, and comes within 1\% of it after a median of 18 to 27.5 simulations; among the baselines, only the evolution strategy does so, after a median of 192 evaluations.
Every decision is recorded, making the search auditable.
\end{abstract}


\maketitle

\section{Introduction}
\label{sec:introduction}

The global energy transition is reshaping how electricity and heat are produced and consumed. At the residential level, distributed energy resources (DERs), including rooftop photovoltaic (PV), battery energy storage systems (BESS), and heat pumps, are expanding rapidly. The International Energy Agency projects continued growth in PV capacity through 2035 under stated policy scenarios \cite{iea2025world}.

Designing building energy systems is challenging because technology selection, capacity sizing, dispatch strategy, and lifecycle economics are tightly coupled. Capacity choices affect dispatch flexibility, while dispatch assumptions influence economically optimal sizes. The design is therefore optimized with respect to the Net Present Cost (NPC), which integrates capital expenditures (CAPEX), replacement costs, fixed and variable operating expenditures (OPEX), and energy purchase costs over the 25-year analysis horizon. 
This problem is a high-dimensional mixed-integer nonlinear optimization task with complex feasibility boundaries. It combines discrete technology inclusion decisions with sizing and performance parameters, and objective evaluation requires a year-long simulation. 
Conventional optimization methods (e.g., mixed-integer programming and evolutionary algorithms) remain valuable but face practical limits in simulation-driven settings with non-convex, multimodal landscapes and expensive black-box evaluations \citep{hafez2012optimal,jung2017optimal}.
Recent advances in Large Language Models (LLMs) suggest an alternative: optimization through iterative reasoning, experimentation, and feedback. Inspired by autonomous experimentation paradigms such as AutoResearch \citep{karpathy2026}, LLM-driven optimization can incorporate domain knowledge, produce explicit, human-readable decision rationales, and adapt strategy over time.
More generally, the object we build is an \emph{agent} in the classical sense of an entity that perceives its environment through sensors and acts upon it through actuators \citep{russell2020aima}: in the LLM instantiation, a reasoning model is augmented with memory, an action interface, and a reflection mechanism, to obtain a goal-directed system \citep{grootendorst2026agents,wang2024survey}.

This paper presents \textit{AutoEnergyDesign}, an LLM-driven framework for building energy system optimization. Our main contributions are: (1) an optimization architecture that couples LLM reasoning with structured experimental feedback, organized around an explicit set of memories, (2) a techno-economic formulation covering eight energy technologies with detailed NPC modeling, (3) a git-based state-management mechanism that preserves best-known configurations while reverting suboptimal updates, and (4) an empirical comparison against random search, a $(1+1)$ evolution strategy with the same acceptance rule as the agent, and Bayesian optimization baselines, using three LLMs of different capability to show how the results depend on the model.

The remainder of this paper is organized as follows. Section \ref{sec:related_works} reviews related literature. Section \ref{sec:methodology} presents the simulator formulation and agent design. Section \ref{sec:case_study} describes the case study, the input data, and the techno-economic assumptions. Section \ref{sec:experiments} describes the experimental protocol and baselines. Section \ref{sec:results} presents results. Section \ref{sec:discussion} discusses implications and limitations. Section \ref{sec:conclusions} concludes the paper.

\section{Related Works}
\label{sec:related_works}

The optimization of building energy systems sits at the intersection of different active research areas: (i) classical methods for energy system design, (ii) emerging applications of Large Language Models (LLMs) in building energy domains, and (iii) LLM-driven optimization paradigms recently developed in the machine learning community. This section reviews each area, identifies key gaps, and positions our contribution within this landscape.

\subsection{Classical Approaches to Energy System Optimization}
\label{subsec:classical}

The design optimization of hybrid residential energy systems has evolved significantly over the past two decades. Early studies established that technology sizing and operational dispatch must be optimized jointly rather than sequentially, particularly for micro-combined heat and power (micro-CHP) and hybrid thermal-electric configurations \cite{Hawkes2007,Merkel2015}.
From a methodological point of view, three primary approaches dominate the literature. \textbf{Mixed-integer programming (MIP)} provides exact solutions for convex approximations of the design problem \cite{hafez2012optimal,jung2017optimal}, but the nonlinearities of component performance curves, economies of scale in costing and discrete technology choices often make it computationally prohibitive for realistic instances with year-long simulation horizons. \textbf{Metaheuristics}, particularly NSGA-II and its variants, remain widely adopted for building and microgrid optimization because they handle nonlinearity, constraints and multiple objectives without requiring gradients \cite{deb2002fast,brownlee2015constrained,Araoye2024,Zhu2021}; while effective for multimodal landscapes, they typically require extensive hyperparameter tuning and large evaluation budgets. For expensive black-box objectives, \textbf{Bayesian optimization} offers superior sample efficiency by fitting a probabilistic surrogate model (typically a Gaussian Process) and optimizing an acquisition function \cite{Snoek2012}; the Tree-structured Parzen Estimator (TPE) \cite{Bergstra2011,akiba2019optuna,watanabe2023tree} extends it to mixed-variable spaces and MOTPE to multiple objectives \cite{ozaki2022multiobjective}, and TPE has become a standard baseline when evaluations are costly.
Despite their successes, these classical methods face persistent limitations in practice: they offer limited traceability regarding why specific configurations are preferred; they struggle to incorporate qualitative domain knowledge without explicit mathematical formulation; and they typically handle infeasible designs through penalty functions or repair operators that must be tailored to each problem.

\subsection{Large Language Models in Building Energy Applications}
\label{subsec:llm_building}

The application of LLMs to building energy systems has expanded rapidly in recent years \citep{liu2025, zhang2025_review, powersys2026survey}. Within this broader landscape, one recurring line of work uses LLMs to support the generation of building energy models from textual descriptions and other early-stage design inputs \citep{jiang2024epluslllm,zhang2025agenticbem,bem2025multiagent}. The main contribution of these studies is a reduction in modeling effort and setup time, rather than the solution of the design optimization problem itself.
Lu et al.~\citep{bem2025multiagent} cover the whole modeling workflow: GPT-4-based agents extract data from CAD drawings and design documents, build the OpenStudio model and configure a genetic algorithm that calibrates it. They do not search the design space.
For operational applications, recent studies have explored LLM-assisted decision-making for economic dispatch, home energy management, and related control tasks in energy systems \citep{Mohammadi2025,ElMakroum2026, jonah2026hems}. An interesting insight is that the LLM performs best when it does not act as the optimizer itself, but drives an exact solver through a structured tool interface.
Multi-agent frameworks have emerged as a particularly active direction. Zhang et al. \cite{Zhang2025} proposed an LLM-powered multi-agent system for power grid control. Madani et al. \cite{Madani2025} surveyed LLM integration in smart grids, highlighting applications in operational decision support. Xia et al. \cite{Xia2026} presented a low-carbon scheduling approach combining multi-agent reinforcement learning with an embedded LLM.
Closer to planning tasks, Forootani et al. \citep{forootani2025rellm} integrate scenario-based energy system optimization with machine-learning surrogates and use an LLM to turn the resulting sensitivity analysis into stakeholder-readable narratives, an approach that shares our traceability motivation but leaves the search itself to a conventional optimizer. Jin et al. \citep{jin2024autoformalism} take the complementary autoformalisation route, using the LLM to translate a natural-language energy-management request into a formal program that is then solved exactly.
Desai~\citep{desai2026agent} uses a single Gemini~1.5~Pro agent to recommend design changes for solar microgrids. A hybrid algorithm refines these suggestions numerically over three variables (number of panels, battery capacity and inverter power), keeping improvements and reverting otherwise. 
Most of these studies address operational control or model generation. The use of LLMs as optimization agents for technology selection and sizing has received less attention; the closest example is \citep{desai2026agent}, with a design vector of three variables.

\subsection{LLM-Driven Optimization Paradigms}
\label{subsec:llm_optimizers}

A distinct but related research stream, originating in the machine learning community, uses LLMs as optimizers: the model proposes candidate solutions on the basis of a problem description and the history of previous evaluations. One influential example is OPRO \cite{yang2024}, where the model receives prior solutions and their scores and generates revised candidates. Although developed for general optimization and prompt-design tasks rather than for engineering systems with physical constraints, it helped establish the idea that language models can participate directly in search.
LEO~\citep{brahmachary2025leo} uses an LLM, mainly GPT-3.5 Turbo, to generate candidates in a population-based optimizer with separate explore and exploit prompts and elitism. LEO comes close to CMA-ES and L-BFGS-B on two-dimensional benchmarks, but for a fixed number of iterations its accuracy decreases as the number of variables grows.
A second line of work has pushed the idea of LLMs as direct optimizers toward more autonomous experimentation. Karpathy's AutoResearch \cite{karpathy2026} showed how an LLM agent can modify code, run experiments, interpret outcomes, and iterate while relying on rollback mechanisms to maintain stability. In a different but related direction, FunSearch \cite{romeraparedes2024} combined language models with evolutionary search to generate and evaluate program variants, showing that LLMs can contribute within a systematic propose-and-test loop. 
The same idea has been employed in \emph{LLM-driven algorithm design}, in which the model evolves the heuristic rather than the solution. In EoH \citep{liu2024eoh}, natural-language ``thoughts'' are coupled with the corresponding code in an evolutionary loop. ReEvo \citep{ye2024reevo} adds an explicit reflection step that verbalises why one heuristic outperformed another before generating the next variant. LLaMEA \citep{vanstein2024llamea} generates and refines entire metaheuristics from benchmark feedback. These developments are surveyed comprehensively in \citep{liu2025llm4ad}. Most of this literature, however, concerns prompt optimization, program synthesis or machine learning workflows, and only rarely engineering design problems with explicit techno-economic constraints, which is the gap that motivates the present work.
A recent survey of LLMs as optimizers \citep{peran2026llmopt} distinguishes \emph{direct} optimization, in which the model itself generates and refines candidate solutions, \emph{tool-augmented} optimization, in which it formalises the problem for an exact solver, and \emph{tool-creating} optimization, in which it synthesises reusable heuristics; AutoEnergyDesign belongs to the first paradigm.
Beyond these foundational works, several recent contributions use LLMs as optimization engines in ways that are directly comparable to ours.
Liu et al.\ \cite{liu2024llambo} proposed LLAMBO, which frames Bayesian optimization (BO) in natural language and uses an LLM to enhance individual BO components without any fine-tuning. LLAMBO is validated on hyperparameter tuning.
AhmadiTeshnizi et al.\ \cite{ahmaditeshnizi2024optimus} introduced OptiMUS, an LLM agent that \emph{formulates and solves} mixed-integer linear programs from natural-language descriptions by writing the mathematical model, generating and debugging solver code, and validating solutions. OptiMUS lowers the barrier to operations-research modeling but delegates the actual optimization to an exact solver, while AutoEnergyDesign keeps the techno-economic model fixed and uses the LLM to propose \emph{configurations} over a simulation-based black box; an exact solver is used only for the inner hourly dispatch.
Liu and Dong~\citep{liu2026hub} let an LLM propose hub capacities from a textual business context, while a linear program evaluates the routing. The LLM itself decides whether to keep each solution, and in one iteration it discards a solution that is better under the hidden true cost model.
AlphaEvolve \cite{novikov2025alphaevolve} orchestrates an ensemble of LLMs in a population-based evolutionary loop that edits and evolves entire codebases, achieving state-of-the-art results on algorithm discovery and infrastructure optimization tasks under substantial compute. Our framework targets a different regime: a single agent under a tight evaluation budget, a sequential keep/revert trajectory rather than a large evolutionary population, and a fixed engineering artifact rather than open-ended algorithm discovery, with emphasis on auditable reasoning traces and explicit techno-economic domain knowledge.
AUTO~\citep{carreon2025auto} applies an LLM search loop to GPU code optimization. A Strategist agent chooses among four strategies (innovate, combine, refine, re-attempt a failure), and several Implementor agents write the code, which is then validated and scored automatically. The Strategist sees a curated subset of past designs: the best, the worst, the most recent and some failed ones. On KernelBench the authors frequently observed cheating, such as caching outputs or hiding the computation from the profiler.
Eureka \cite{ma2024eureka} uses a coding LLM to evolve \emph{reward functions} for reinforcement learning, with environment code as context and a reward-reflection step, evaluating each candidate through full RL training runs. The LLM output is the code that \emph{defines} the objective, while AutoEnergyDesign optimizes the decision variables directly against a fixed economic objective evaluated by a deterministic simulator.
Taken together, these works show that LLMs can act as components of BO (LLAMBO), as formulators for exact solvers (OptiMUS), as candidate generators (LEO), as strategists of code search (AUTO), as evolutionary code optimizers (AlphaEvolve, EoH, ReEvo, LLaMEA), as proposers of design decisions evaluated by a solver (Liu and Dong), or as designers of objectives (Eureka). AutoEnergyDesign occupies a complementary point in this space: a single, budget-constrained, auditable agent that performs the search \emph{directly} over a constrained techno-economic design space, with persistent episodic and reflective memories (Table~\ref{tab:agent_files}) and git-based rollback as a state-management layer enforced by the environment rather than left to the model.

\subsection{Identified Gaps and Research Opportunities}
\label{subsec:gaps}

From this literature review, we identify several key gaps that motivate our work:

\textbf{Gap 1: Traceability of optimization decisions.} Classical black-box optimizers (Bayesian, evolutionary) return solutions with minimal explanation. Practitioners cannot easily understand \textit{why} a particular configuration is preferred. LLM-driven methods produce reasoning traces, such as the decision table of \citep{liu2026hub}, but these traces can be wrong and need to be checked against the outcome of each evaluation.

\textbf{Gap 2: Integration of qualitative domain knowledge.} Engineering design relies heavily on heuristics, physical intuition, and economic insights that are difficult to encode mathematically. Existing optimizers require explicit formulation of such knowledge as constraints or penalty functions. LLMs can absorb qualitative knowledge through prompting, but frameworks for systematic integration with simulation-based optimization are still few.

\textbf{Gap 3: Accessibility for practitioners.} Advanced optimization tools require expertise in mathematical programming, algorithm configuration, or machine learning. There is a need for optimization workflows that are accessible to energy engineers without requiring deep optimization expertise.

\subsection{Positioning of This Work}
\label{subsec:positioning}

AutoEnergyDesign addresses these gaps through the following features of the framework, which correspond to contributions (1) and (3) of Section~\ref{sec:introduction}:

\begin{enumerate}
    \item \textbf{Traceable and auditable optimization.} Persistent reasoning traces (agent notes, and a decision log in which every experiment is justified before it is run) let practitioners audit the optimization trajectory and check each rationale against the outcome logged by the environment, while the natural-language interface lowers the barrier to applying advanced optimization to energy system design.

    \item \textbf{Structured domain knowledge integration.} Operational instructions are separated from domain knowledge (\texttt{insights.md}), so that engineers can inject qualitative heuristics without modifying code or mathematical formulations; \emph{procedural} memory (how to conduct the search) and \emph{semantic} memory (what is true about the techno-economic context) can thus be edited and audited independently \citep{sumers2024coala}.

    \item \textbf{Git-based safe exploration.} The commit/revert mechanism never loses a known-good configuration, which allows aggressive exploration, extending the version-control concept of AutoResearch to simulation-driven engineering optimization.
\end{enumerate}

The closest design-phase work \citep{desai2026agent} optimizes three variables with a human-in-the-loop agent. The present work selects and sizes eight technologies against a 25-year NPC, with an autonomous loop compared with black-box optimizers over repeated runs.
The purpose of this study is not to replace exact optimization methods when a suitable mathematical formulation is available. If both equipment sizing and operational dispatch can be represented by linear relationships and integer decisions, they can be optimized jointly using mixed-integer linear programming (MILP). In that setting, an exact solver can provide optimality guarantees that the proposed LLM agent does not offer.
Our framework instead targets simulation-based design problems that are difficult to express in this form, for example because they involve nonlinear component models, tabulated performance data, economies of scale, or an existing simulator whose internal formulation cannot be modified. In these settings, the agent proposes designs and learns from their simulated performance without requiring an explicit mathematical formulation of the entire design problem. The present case study uses linear-programming dispatch to provide a transparent and well-defined evaluation environment; it demonstrates the agent's search behavior in a controlled setting, rather than establishing that exact optimization is infeasible for this instance or that the agent is superior to an exact solver.

\section{Methodology}
\label{sec:methodology}

This section describes the LLM-driven agent and then the energy system simulator that evaluates its proposals.

\subsection{Agent Architecture}
\label{subsec:agent}

\begin{figure}[pos=htbp]
\centering
\includegraphics[width=0.8\textwidth]{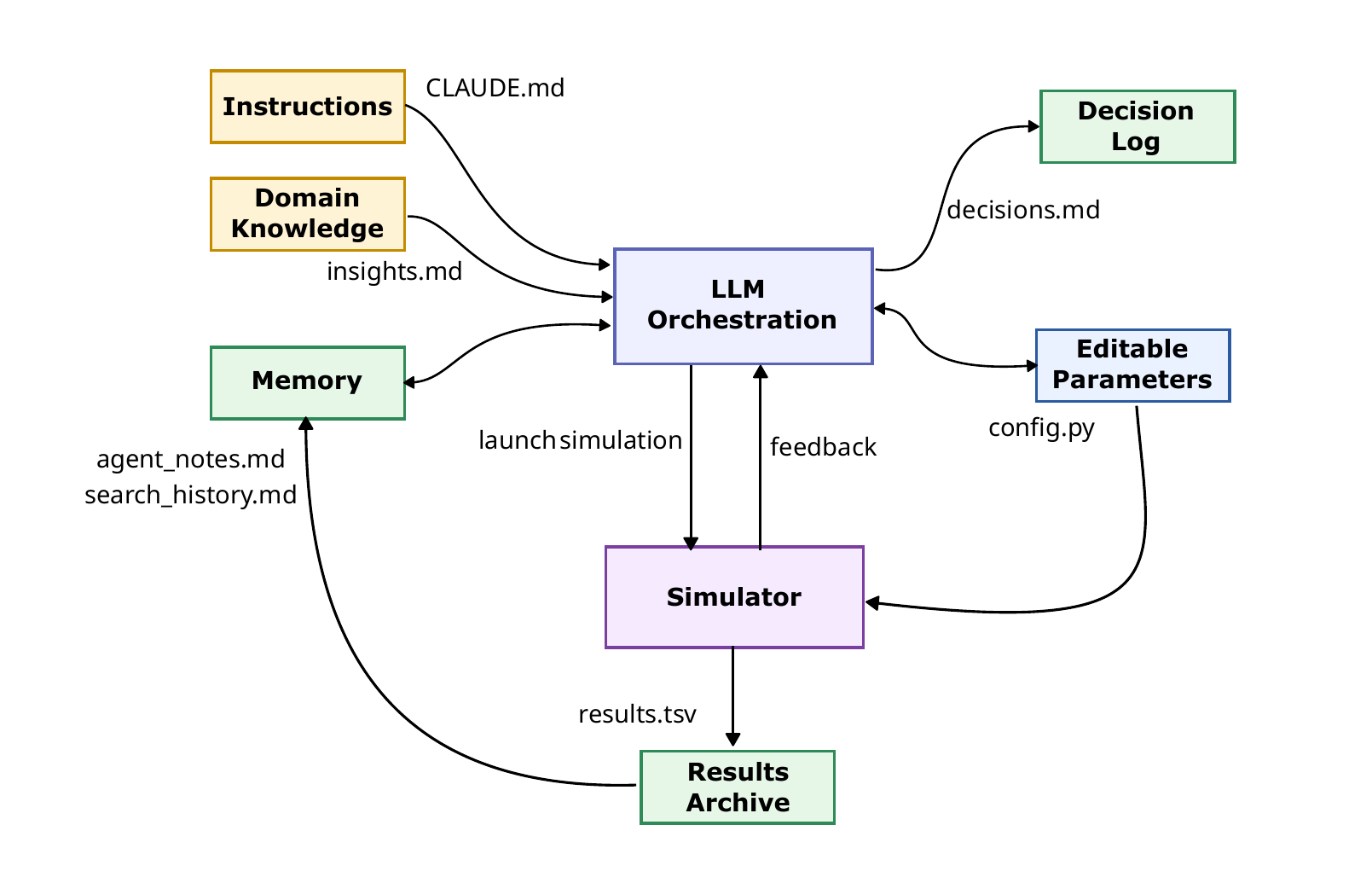}
\caption{Conceptual scheme of the AutoEnergyDesign agent: instruction and domain files, editable configuration, simulator, hooks and logs.}
\label{fig:file_workflow}
\end{figure}

The agent implements an iterative experimentation loop inspired by Karpathy's AutoResearch \cite{karpathy2026}: a language model repeatedly edits a configuration file, runs the simulator, and decides the next change on the basis of the results obtained so far. The model is not called through an API by a custom runner but operates inside a general-purpose coding agent, Claude Code, which gives it a file reader, a file editor and a shell. Everything else (the simulator, the validator, the acceptance rule and the logs) consists of ordinary scripts and files. The files and components involved are shown schematically in Figure~\ref{fig:file_workflow}. 

\subsubsection{Overview of the Loop}

Each run is a single coding-agent session started in the project directory with the message ``Run CLAUDE.md''.\footnote{\texttt{claude ---model <model> ---session-id <id> ---dangerously-skip-permissions ---max-turns 1000 "Run CLAUDE.md"}. Permission prompts are disabled so that the session can run unattended.} Before the session starts, the launch script simulates the default configuration and records it as experiment~1; the agent then performs experiments 2 to 51, i.e.\ a budget of $B=50$. Each experiment follows the same procedure, prescribed by the instruction file:
\begin{enumerate}[noitemsep]
    \item read the state files (Table~\ref{tab:agent_files}) and record the outcome of the previous experiment in the decision log;
    \item stop if the experiment counter has reached 51;
    \item every $N_a=4$ experiments, rewrite the reflective note \texttt{agent\_notes.md};
    \item record the next decision in \texttt{decisions.md};
    \item edit \texttt{config.py};
    \item run \texttt{python simulate.py}, then return to step~1.
\end{enumerate}
Two hooks of the coding agent, triggered before and after every shell command, enforce the rules independently of the model. The pre-execution hook validates \texttt{config.py} whenever the command launches the simulator (admissible range, quantization step, at least one thermal source, inverter flags tied to their parent technology) and blocks invalid proposals before the simulation is launched, so that they do not consume an evaluation. The post-execution hook detects that a new simulation has been run, commits the configuration, applies the acceptance rule of Eq.~\eqref{eq:accept}, restores the previous configuration if the change is rejected, and updates the logs and the state files. 
The agent observes the result of each experiment in two ways. The simulator output, returned as the result of the shell command, reports the NPC and its breakdown (CAPEX by technology, salvage credit, fixed and variable OPEX, electricity and gas costs), the unserved electricity, heat and cooling, the CO$_2$ emissions and the self-sufficiency. The status file written by the hook reports the verdict (kept, worse, infeasible, out of range or crashed). 

\subsubsection{Files and Memory}

Table~\ref{tab:agent_files} lists the files involved in the loop, classified according to the memory types of the CoALA framework \citep{sumers2024coala}. The main design choice is the separation between what the environment writes, which is reliable, and what the agent writes, which records its reasoning and may be wrong.
\begin{table}[pos=htbp]
\centering
\caption{Files of the agent loop. ``Reads'' indicates whether the agent is instructed to read the file at every experiment.}
\label{tab:agent_files}
\small
\setlength{\tabcolsep}{4pt}
\begin{tabular}{@{}>{\raggedright\arraybackslash}p{0.25\textwidth} >{\raggedright\arraybackslash}p{0.13\textwidth} >{\raggedright\arraybackslash}p{0.10\textwidth} c >{\raggedright\arraybackslash}p{0.33\textwidth}@{}}
\toprule
File & Memory & Written by & Reads & Content \\
\midrule
\texttt{CLAUDE.md} & procedural & authors & yes & Procedure and rules (Section~\ref{subsubsec:instructions}) \\
\texttt{insights.md} & semantic & authors & yes & Energy prices; per technology: efficiency, CAPEX, OPEX, lifetime; replacement and salvage rules \\
\texttt{config.py} & -- & agent & yes & Current design, with admissible range and step of each variable in comments \\
\texttt{.exp\_counter}, \texttt{.last\_status} & episodic & hook & yes & Experiment counter; verdict of the last experiment \\
\texttt{search\_history.md} & episodic & hook & yes & All kept designs ranked by NPC and the $K=10$ most recent discarded ones, with reason and difference from the best \\
\texttt{decisions.md} & episodic & agent & yes & One entry per experiment: change, rationale, expected effect, alternatives considered, observed outcome \\
\texttt{agent\_notes.md} & reflective & agent & yes & Note of at most ten lines, rewritten every $N_a=4$ experiments \\
\texttt{results.tsv}, git history & -- & hook & no & Complete log of all trials with parameters and outcome \\
\bottomrule
\end{tabular}
\end{table}
Working memory is the context of the session itself, which accumulates the files read, the simulation outputs and the agent's own messages; the state files are re-read at every experiment so that the procedure does not depend on what survives a compaction of the context. The search history is a compact but lossy summary of the complete log: designs are described by the sizes of the enabled technologies only, without efficiencies, inverter ratings or battery power.
The reflective note follows a fixed template with four headings: \emph{Worked} (changes that reduced the NPC, with direction and amount), \emph{Failed} (changes that were worse, infeasible or out of range), \emph{Behavior} (sensitive technologies, interactions, exhausted regions) and \emph{Next} (two or three experiments to try). It condenses the trajectory into verbal lessons, in the spirit of Reflexion \citep{shinn2023reflexion}, and its \emph{Next} section is the starting point for the following decisions. The decision log complements it at the level of single steps: since each expectation is written before the result is known, every prediction can be checked afterwards against the outcome, and the log is the main source for the qualitative analysis of Appendix~\ref{app:qualitative}.

\subsubsection{Instructions and Domain Knowledge}
\label{subsubsec:instructions}

The instruction file \texttt{CLAUDE.md} contains the per-experiment procedure above and the following rules: (i)~minimize the NPC reported by the simulator; (ii)~edit only \texttt{config.py}, \texttt{agent\_notes.md} and \texttt{decisions.md}, and do not read or modify the simulator, the cost parameters or the hooks; (iii)~respect the range and step of each variable, keep at least one thermal source (boiler, heat pump or CHP) enabled, and leave the inverter activation lines unchanged; (iv)~change one to three logically related parameters per experiment, and in the first experiments switch technologies on and off before fine-tuning sizes; (v)~after a rejected experiment, do not retry the same values; (vi)~never ask questions and never stop before the budget is exhausted.
The domain-knowledge file \texttt{insights.md} contains only input data, copied from the cost parameters of the simulator: energy prices (electricity purchase 0.30~EUR/kWh, export 0.08~EUR/kWh, natural gas 0.12~EUR/kWh); for each technology, its efficiency or performance data, specific CAPEX and OPEX and lifetime; the time-shared operation of the reversible heat pump, stated as a rule (heating share plus cooling share at most 100\% of its size in each hour); the fact that inverter costs are included in those of PV, wind and battery; the absence of thermal storage; the replacement and salvage rules; and the absence of active self-sufficiency or CO$_2$ constraints. It contains no rankings of technologies and no design guidance. Since the agent is told not to read the simulator or its parameters, this file is its only source of information on the scenario besides the simulator output.
The rules on files are a convention rather than a sandbox: with permission prompts disabled, the agent could in principle read or modify any file, and compliance is checked afterwards on the session transcripts and the git history of each run. The design space, by contrast, is enforced by the pre-execution hook regardless of the model's behaviour.

\subsubsection{Acceptance Rule and State Management}

At experiment $k$, the proposed configuration $\mathbf{x}_k$ is compared with $\mathbf{x}^\star_{k-1}$, the best valid configuration found so far. The best configuration is updated only if the new design is valid (feasible, Section~\ref{subsubsec:metrics}) and has a strictly lower cost:
\begin{equation}
\mathbf{x}^\star_{k} =
\begin{cases}
\mathbf{x}_k, & \text{if } \mathbf{x}_k \text{ is valid and } \mathrm{NPC}(\mathbf{x}_k) < \mathrm{NPC}(\mathbf{x}^\star_{k-1}),\\[2pt]
\mathbf{x}^\star_{k-1}, & \text{otherwise,}
\end{cases}
\label{eq:accept}
\end{equation}
with $\mathrm{NPC}(\mathbf{x}^\star_0)=+\infty$, so that the first feasible design is always kept; this matters because the default configuration is infeasible.

The rule is implemented with git. Each run works on a dedicated branch, \texttt{exp/<timestamp>}. After every simulation the post-execution hook commits \texttt{config.py}; if the design is kept, it records the new best NPC, and otherwise it restores the previous \texttt{config.py} and commits the revert. 
Every trial, kept or discarded, is appended to \texttt{results.tsv} with its full parameter vector, feasibility flag and outcome, and \texttt{search\_history.md} and \texttt{.last\_status} are regenerated from it. The branch therefore contains the complete, reproducible trajectory of the run, and the working copy always holds the best design found so far.

\subsection{Simulator}
\label{subsec:simulator}

\begin{figure}[pos=htbp]
\centering
\includegraphics[width=0.7\textwidth]{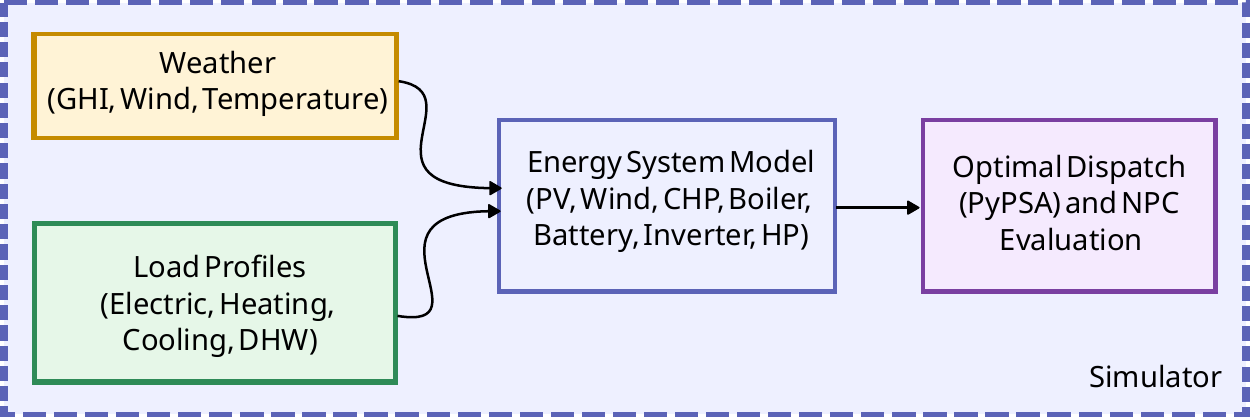}
\caption{Conceptual Scheme of the Simulator}
\label{fig:simulator}
\end{figure}

The simulator forms the core of the optimization framework, evaluating candidate energy system configurations through year-long operational simulation and techno-economic assessment. The simulator is implemented using PyPSA (Python for Power System Analysis) \citep{Brown2018}, an open-source framework for power system modeling and optimization. PyPSA is used here as an operational solver. Figure~\ref{fig:simulator} shows a schematic of the simulated system.

\subsubsection{Problem Formulation}

The energy system design optimization problem can be stated as follows. Given a set of available technologies $\mathcal{T}$, hourly electricity, heat and cooling demands over a planning horizon $t=1,\dots,T$, and economic parameters such as energy prices and discount rate, the objective is to identify the design vector $\mathbf{x}$ (technology activation and sizing) and the operational dispatch $\mathbf{u}_{1:T}$ that minimize lifecycle Net Present Cost while meeting demand at each time step. Demand that the system cannot meet is represented explicitly as unserved energy, which is penalised in the NPC.

Mathematically, we write:

\begin{align}
    \min_{\mathbf{x},\,\mathbf{u}_{1:T}} \quad & \text{NPC}(\mathbf{x},\mathbf{u}_{1:T}) \\
    \text{s.t.} \quad & \mathbf{x} \in \mathcal{X} && \text{(design bounds / discrete choices)} \\
    & g(\mathbf{x},\mathbf{u}_{1:T}) \le 0 && \text{(technical/operational constraints)} \\
    & p_t^{\mathrm{sup}}(\mathbf{x},\mathbf{u}_t) + s_t^{\mathrm{el}} = d_t^{\mathrm{el}}\;\; \forall t && \text{(electricity balance)} \\
    & q_t^{\mathrm{heat}}(\mathbf{x},\mathbf{u}_t) + s_t^{\mathrm{heat}} = d_t^{\mathrm{heat}}\;\; \forall t && \text{(heat balance)}\\
    & q_t^{\mathrm{cool}}(\mathbf{x},\mathbf{u}_t) + s_t^{\mathrm{cool}} = d_t^{\mathrm{cool}}\;\; \forall t && \text{(cooling balance)} \\
    & s_t^{\mathrm{el}},\ s_t^{\mathrm{heat}},\ s_t^{\mathrm{cool}} \ge 0\;\; \forall t && \text{(unserved energy)}
\end{align}

where $\mathbf{x}$ is the vector of design parameters (technology enablement and sizing), and $\mathbf{u}_{1:T}=\{\mathbf{u}_t\}_{t=1}^T$ is the sequence of hourly operational decisions (dispatch, charging/discharging, imports/exports). Here, $d_t^{\mathrm{el}}$, $d_t^{\mathrm{heat}}$ and $d_t^{\mathrm{cool}}$ denote the exogenous electricity, heat and cooling demands at hour $t$, while $p_t^{\mathrm{sup}}$, $q_t^{\mathrm{heat}}$ and $q_t^{\mathrm{cool}}$ denote the electricity, heat and cooling supplied to demand by the system, and $s_t^{\mathrm{el}}$, $s_t^{\mathrm{heat}}$ and $s_t^{\mathrm{cool}}$ the electricity, heat and cooling demand that is not served (written $s^{\mathrm{el}}(t)$, $s^{\mathrm{heat}}(t)$, $s^{\mathrm{cool}}(t)$ in Section~\ref{sec:dispatch}).
The set $\mathcal{X}$ encodes admissible design choices, and $g(\mathbf{x},\mathbf{u}_{1:T})\le 0$ collects technical-operational constraints (component limits, storage dynamics, conversion efficiencies, and network balances).
Each kilowatt-hour of unserved energy is charged at a penalty $\pi = 1000$~EUR/kWh, which enters the NPC through the variable operating cost (Section~\ref{sec:dispatch}); feasibility is defined in Section~\ref{subsubsec:metrics}.
Each inverter is a dedicated technology: it has its own rated power and efficiency, but its activation is not an independent decision, it is tied to the activation of the technology it serves (PV, battery, or wind). 
The heat pump is a single \emph{reversible} machine that provides space heating and space cooling.
The design parameter vector $\mathbf{x}$ comprises the variables reported in Table~\ref{tab:design_variables}, whose definitions and admissible ranges (Table~\ref{tab:design_bounds}) are given in Appendix~\ref{app:params}.
The search exposes seven freely settable binary activation variables: $z_{\mathrm{pv}}$, $z_{\mathrm{wt}}$, $z_{\mathrm{chp}}$, $z_{\mathrm{st}}$, $z_{\mathrm{boi}}$, $z_{\mathrm{bat}}$, and $z_{\mathrm{hp}}$. The three inverter activations are constrained to the corresponding technologies, and the grid is always available. 
The symbols are summarized in Appendix~\ref{app:nomen}

\subsubsection{Technology Models}

The simulator includes eight energy technologies, whose models are described in Appendix~\ref{app:tech}.

\subsubsection{Net Present Cost Calculation}

The objective function is the Net Present Cost (NPC) of lifecycle costs, computed as the sum of discounted costs over the 25-year analysis horizon with a discount rate $r=4\%$:
\begin{equation}
    \text{NPC} = \text{CAPEX}_{\text{total}} - \textstyle\sum_i S_i + \text{OPEX}_{\text{fixed}}^{\text{PV}} + \text{OPEX}_{\text{variable}} \times \text{PV}_{\text{factor}}
\label{eq:npc}
\end{equation}
where $\text{CAPEX}_{\text{total}}$ is the discounted investment, including the replacement of the components whose lifetime is shorter than the horizon, $S_i$ is the residual value credited for the equipment of technology $i$ still usable at the end of the horizon, $\text{OPEX}_{\text{fixed}}^{\text{PV}}$ is the present value of the fixed operating costs, $\text{OPEX}_{\text{variable}}$ is the annual variable operating cost given by the optimal dispatch (Section~\ref{sec:dispatch}), including the penalty on unserved energy, and $\text{PV}_{\text{factor}} \approx 15.62$ is the annuity present-value factor. No decommissioning or end-of-life cost is included. The detailed expressions are given in Appendix~\ref{app:npc}.
At the methodological level, the formulation remains generic: component costs, lifetimes, replacement schedules, and fixed operating costs are treated as input parameters that can be adapted to the application context. The specific numerical values adopted in this paper are introduced later in the case study through Tables \ref{tab:capex} and \ref{tab:opex} (Appendix~\ref{app:params}). These assumptions instantiate one evaluation scenario, but they do not limit the generality of the proposed optimization framework.
The variable operating cost of the first simulated year is assumed to repeat, in real terms, for all 25 years; no load growth, no efficiency degradation and no tariff evolution are represented.

\subsubsection{Optimal Dispatch Formulation}
\label{sec:dispatch}

The operational dispatch is formulated as a linear program, solved over all hours of the year, that minimizes the total variable operating cost subject to demand satisfaction and equipment constraints.
The complete formulation (decision variables, objective function and constraints) is given in Appendix~\ref{app:dispatch}.
The heat-demand term $p_{\text{load}}^{\text{heat}}$ denotes space heating; domestic hot water $p_{\text{load}}^{\text{dhw}}$ is added to the same heat bus. The load terms coincide with the demands of the problem formulation: $p_{\text{load}}^{\text{el}}(t)=d_t^{\mathrm{el}}$, $p_{\text{load}}^{\text{heat}}(t)+p_{\text{load}}^{\text{dhw}}(t)=d_t^{\mathrm{heat}}$ and $p_{\text{load}}^{\text{cool}}(t)=d_t^{\mathrm{cool}}$. The LP is solved with HiGHS through PyPSA with a time limit of 600~s; a solve that does not terminate with an optimal status is treated as a crash.

\subsubsection{Reported Metrics and Feasibility}
\label{subsubsec:metrics}

Beyond the NPC objective, each simulation reports two secondary indicators that are logged but not optimized. The annual carbon footprint aggregates grid imports and fuel combustion using emission factors of $0.400~\mathrm{kgCO_2/kWh}$ for grid electricity and $0.202~\mathrm{kgCO_2/kWh}$ for natural gas (boiler and CHP). Electrical self-sufficiency is reported as $1 - E_{\mathrm{grid}}^{\mathrm{import}} / E_{\mathrm{el}}^{\mathrm{tot}}$, where $E_{\mathrm{el}}^{\mathrm{tot}}$ is the total electrical consumption including heat-pump operation in both heating and cooling modes.

\paragraph{Definition of feasibility.} The unserved-energy variables make the dispatch linear program solvable for every design. A trial is discarded when at least one of the following holds: (1)~a design variable falls outside its admissible range, or off its quantization grid, or no thermal source is enabled, etc. (label \textsc{out\_of\_range}, detected before the solve by the pre-execution hook, and checked again by the post-execution hook as a safeguard); (2)~the simulation raises an exception or the solver returns a non-optimal status (label \textsc{crash}); (3)~the optimal dispatch exhibits non-zero unserved energy, $\sum_t (s^{\text{el}}(t)+s^{\text{heat}}(t)+s^{\text{cool}}(t))>\varepsilon$ with $\varepsilon=10^{-3}$~kWh (label \textsc{infeasible}); (4)~the design is feasible but its NPC is not lower than that of the best current design (label \textsc{worse}).
The \emph{feasible rate} of Table~\ref{tab:multi_seed} counts only criterion~3: it is the fraction of evaluated configurations whose optimal dispatch serves the whole demand. Configurations failing criterion~1 are rejected by the pre-execution validator, never reach the simulator and are not counted as evaluations.

\section{Case Study}
\label{sec:case_study}

The case study evaluates the AutoEnergyDesign framework using measured load profiles representative of an office building in Southern Italy (Naples $\approx 40.9^{\circ}$N, $14.3^{\circ}$E), operated on weekdays with space heating/cooling (HVAC) active 07:00--18:00, including early-morning pre-conditioning, and domestic hot water drawn over 08:00--17:00.

\subsection{Scenario Assumptions}

For the scenario examined here, the electricity purchase price is set to 0.30 EUR/kWh and the natural-gas price to 0.12 EUR/kWh. In addition, the export tariff is set to $c_{\text{export}} = 0.08$ EUR/kWh.
Both energy prices are treated as flat, time-invariant tariffs. This is a simplification: a flat electricity price removes any arbitrage value from the battery, which can then only be justified by self-consumption of on-site generation, i.e.\ by the 0.22~EUR/kWh spread between purchase and export prices.
Table~\ref{tab:capex} in Appendix~\ref{app:params} summarizes the capital-cost, lifetime, and replacement assumptions adopted in the case study \citep{DEA2026}.
Table~\ref{tab:opex} in Appendix~\ref{app:params} reports the fixed OPEX assumptions used in the considered scenario \citep{DEA2026}.

\subsection{Consumption and Weather Data}

The profiles used in the experiments are not synthetic but are constructed from measured hourly electricity consumption and hourly weather observations for the year 2024 (8,784 hourly samples, including 29 February). The weather series (solar irradiance, wind speed at 10~m, air temperature) are obtained from the Open-Meteo reanalysis for the coordinates of the site. The measured series is an electrical load that also includes the electricity used for space heating, space cooling and domestic hot water (DHW); since these end uses are not metered separately, the split is estimated from heating and cooling degree-hours and occupancy. Over the year the resulting demands are 703~MWh of electricity, 459~MWh of space heating, 277~MWh of cooling (peak 664~kW\textsubscript{th}) and 215~MWh of DHW. Space heating and cooling never occur in the same hour, whereas DHW can coincide with cooling. The list of series, their statistics (Table~\ref{tab:profile_stats}), the annual profiles (Figure~\ref{fig:profiles}) and the details of the split are given in Appendix~\ref{app:data}.

\section{Experiments}
\label{sec:experiments}

\subsection{Experimental Setup}

All methods share the same simulation environment, objective definition and feasibility handling, so that differences in performance reflect the search strategy rather than the surrounding infrastructure. The agent is run for a maximum of $B=50$ experiments, with ten independent runs per model, whereas each baseline is run for $300$ evaluations per seed over ten seeds.
The agent is driven by the Claude Code coding agent (version 2.1.296) with three Claude models of increasing capability: Haiku~5.5, Sonnet~5.5 and Opus~5.5, all with the reasoning effort set to \emph{medium}.
All runs start from the same initial \texttt{config.py}, in which only a heat pump of 500~kW is installed, use the same instruction and domain-knowledge files, and differ only in the model and in the non-determinism of its sampling. This default configuration, simulated as experiment~1 (Section~\ref{subsec:agent}), is infeasible: it leaves part of the heating and cooling demand unserved, and its penalised NPC is about 977~MEUR. It is not counted in the budget of 50 and is excluded from the statistics.
To measure the contribution of the domain-knowledge file, ten further Opus~5.5 runs and ten further Haiku~5.5 runs are performed without access to \texttt{insights.md}: the file remains in the project directory, but is dropped from the list of files the agent reads at every experiment and added to the files the agent must not read, and the decision template no longer mentions it; everything else is unchanged (Section~\ref{subsec:ablation_insights}).
Each run is stopped when the experiment counter reaches the budget or when the session ends. Twenty-nine of the thirty sessions used the whole budget; one Haiku~5.5 session ended after 36 experiments (35 chosen by the agent). All twenty ablation sessions used the whole budget.
For every run we record the best feasible NPC found so far after each evaluation. Each method is then described by the best and the median of the final values over its runs, considering only the feasible solutions, and by the number of runs whose final design is the best one found in the study, whose NPC we denote by $\mathrm{NPC}^{\star}$. 

We also report the number of evaluations needed to come within a relative tolerance $\delta$ of the best design: for each run, this is the first evaluation at which the best feasible NPC found so far is at most $(1+\delta)\,\mathrm{NPC}^{\star}$. 
Runs that never reach it within their budget are counted as not reaching it, and the median and the maximum number of evaluations are computed over the runs that reach it; both numbers are reported together with the fraction of such runs, which measures the reliability of the method. We use $\delta=1\%$.

\subsection{Baselines}

To assess the quality of the proposed approach, we compare it against three baselines that represent complementary forms of black-box search: unguided random exploration, surrogate-guided Bayesian optimization, and a local search that uses the same acceptance rule as the agent.

\paragraph{Random Search}
Random search serves as the simplest reference method, drawing configurations uniformly from the admissible design space without using information from previous evaluations. In the present setting, boolean activation variables are sampled uniformly over their two possible states, while continuous parameters are sampled uniformly on the quantization grid of their admissible ranges (Optuna \texttt{RandomSampler}, with the same parametrisation used for TPE). 
Unlike the agent, random search and TPE are not filtered by the pre-execution validator: designs without a thermal source are simulated, found infeasible, and counted as evaluations. Although conceptually simple, random search remains a meaningful benchmark in high-dimensional optimization because it can explore broad regions of the space without the structural biases introduced by more elaborate methods.

\paragraph{Bayesian Optimization with TPE}
As a stronger baseline, we use Bayesian optimization through the Tree-structured Parzen Estimator (TPE). TPE models the objective distribution by separating past observations into relatively good and relatively bad regions and then selecting new candidates by maximizing expected improvement under this density-based representation. The implementation is based on Optuna \citep{akiba2019optuna} ver. 4.8.0, using its define-by-run framework and standard TPE sampler. 

\paragraph{Local Search with a $(1+1)$ Evolution Strategy}
Because the proposed agent implements a strict hill-climbing acceptance rule (Eq. \eqref{eq:accept}), the natural question is how much of its performance comes from the acceptance rule and how much from the language model choosing the move. To answer it, the third baseline is a $(1+1)$ evolution strategy on the quantized design vector that keeps the acceptance rule and replaces the language model with a random move. The genotype consists of the seven activation flags and of the grid index of each of the 18 continuous variables, so every candidate lies on the admissible grid by construction. The search starts from the same default configuration as the agent, whose evaluation is the first of the 300. Each offspring perturbs between one and three active genes, chosen uniformly (mirroring the one-to-three-parameter rule given to the agent): a flag is flipped, and when a technology is switched on its parameters are resampled uniformly on their grids; an index is shifted by a Gaussian step whose standard deviation is a fraction $\sigma_{\mathrm{ES}}$ of the number of grid levels ($\sigma_{\mathrm{ES},0}=0.1$), adapted with the 1/5 success rule (multiplied by 1.5 after a success and by $1.5^{-1/4}$ after a failure, within $[0.01, 0.5]$). Offspring with no thermal source, or identical to the parent, are redrawn without being simulated. An offspring replaces the parent under Eq.~\eqref{eq:accept}; before the first feasible design is found, it is accepted whenever it lowers the penalised NPC, so that the search can walk towards feasibility from the infeasible start.

\section{Results}
\label{sec:results}

\begin{figure}[pos=htbp]
\centering
\includegraphics[width=1\textwidth]{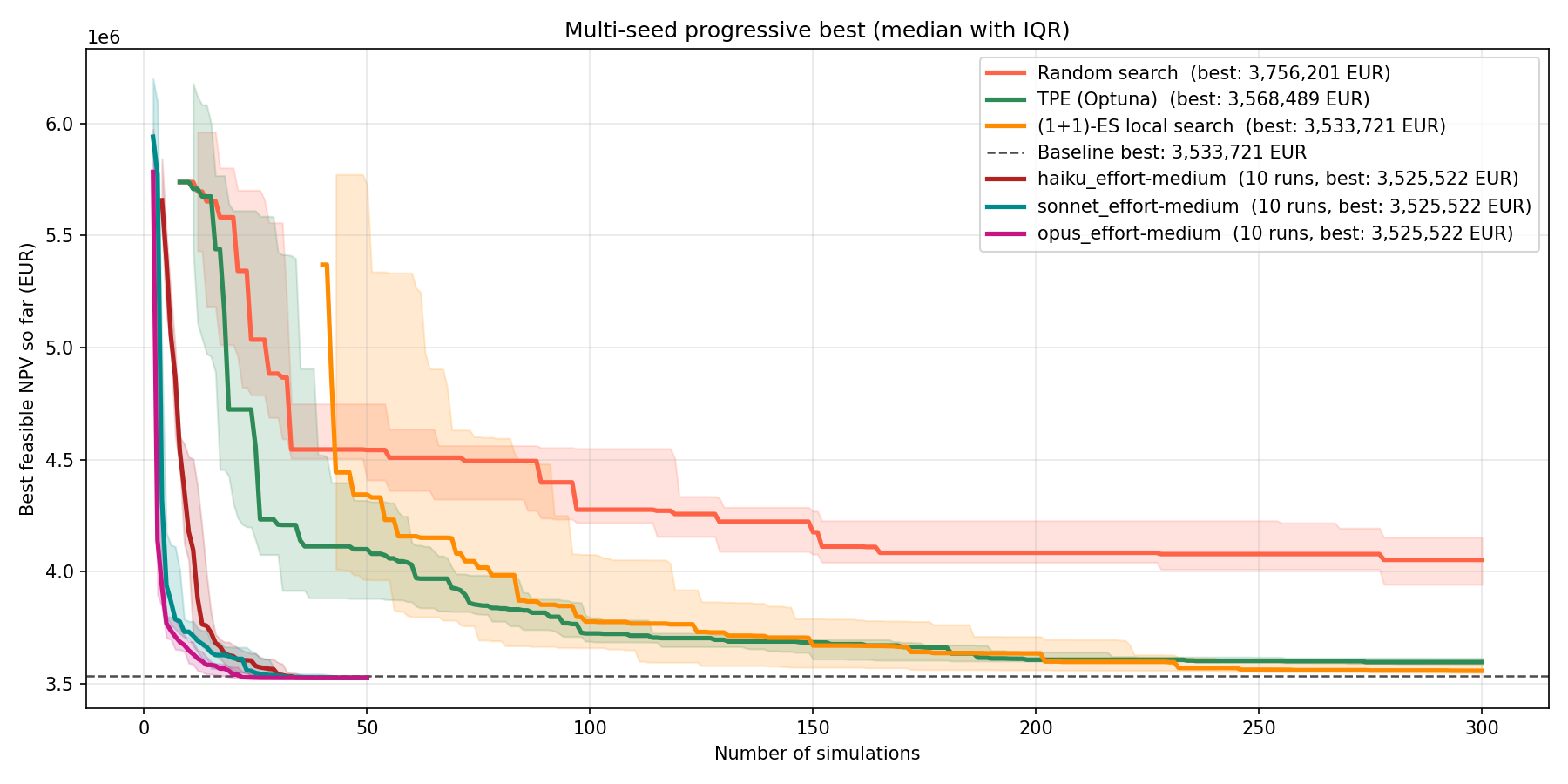}
\caption{Best feasible NPC found within the first $x$ evaluations. At each $x$, the line is the median across runs (ten seeds per baseline, ten runs per agent) and the band is the interquartile range, both computed pointwise over the per-run best-so-far curves; runs with no feasible design yet count as $+\infty$, so some curves start later. Agent curves stop at 50 simulations. The dashed line is the best final baseline value.}
\label{fig:comparison_results}
\end{figure}

Figure~\ref{fig:comparison_results} shows the convergence behavior of the methods as a function of the available evaluation budget, and Table~\ref{tab:multi_seed} summarises the final values and the number of evaluations each method needs to come within 1\% of the best design found in the study (3{,}525{,}522~EUR, i.e.\ at most 3{,}560{,}777~EUR) within their budget. Baseline statistics are computed over ten seeds of 300 evaluations, and agent statistics over ten runs of at most 50 simulations.
The ablation without the domain-knowledge file is reported separately in Section~\ref{subsec:ablation_insights}.

\begin{table}[pos=htbp]
\centering
\caption{Final Net Present Cost (EUR) reached by each method over ten runs.
``Best'' and ``Median'' are the best and median final NPC.
``Feasible rate'' is the fraction of evaluations whose dispatch satisfies the whole demand.
``Evals to 1\% of best'' reports how many runs reach a best feasible NPC within 1\% of the best design found in the study, with the median and, in parentheses, the maximum number of evaluations needed.
``Runs at best'' is the number of runs whose final design is the best one found in the study.}
\label{tab:multi_seed}
\footnotesize
\setlength{\tabcolsep}{4pt}
\begin{tabular}{@{}lccccc@{}}
\toprule
Method & Best & Median & \hd{Feasible\\rate} & \hd{Evals to 1\%\\of best} & \hd{Runs\\at best} \\
\midrule
Random search & 3,756,201 & 4,052,571 & 19.8\% & 0/10 & 0/10 \\
TPE (Optuna) & 3,568,489 & 3,596,245 & 70.1\% & 0/10 & 0/10 \\
$(1+1)$-ES & 3,533,721 & 3,557,264 & 73.3\% & 5/10 $\cdot$ 192 (246) & 0/10 \\
\midrule
Agent, Haiku~5.5$^{a}$ & \textbf{3,525,522} & 3,525,947 & 87.6\% & \textbf{10/10} $\cdot$ 27.5 (45) & 1/10 \\
Agent, Sonnet~5.5 & \textbf{3,525,522} & \textbf{3,525,522} & 89.8\% & 8/10 $\cdot$ 20.5 (30) & 6/10 \\
Agent, Opus~5.5 & \textbf{3,525,522} & \textbf{3,525,522} & 89.0\% & \textbf{10/10} $\cdot$ \textbf{18 (31)} & \textbf{10/10} \\
\bottomrule
\end{tabular}

\vspace{2pt}
\parbox{\linewidth}{\footnotesize $^{a}$ One of the ten Haiku~5.5 sessions ended before the budget, after 35 experiments chosen by the agent.}
\end{table}

\subsection{Baselines}
Random search remains far from the other methods, and only a small fraction of its evaluations satisfy the whole demand (feasible results). TPE substantially improves both the final value and the feasible rate. The $(1+1)$ evolution strategy, which shares the agent's acceptance rule, attains the best baseline results (Table~\ref{tab:multi_seed}), although one of its ten seeds remains trapped at 4.782~MEUR with a PV field of only 560~m$^2$.
No random-search or TPE seed reaches $(1+\delta)\,\mathrm{NPC}^{\star}$ in 300 evaluations; only the $(1+1)$-ES enters this band, in half of its seeds, while the other half never reach it.

\subsection{Agent} All three models reach $\mathrm{NPC}^{\star}=$ 3{,}525{,}522~EUR, the best value found by any method, in at least one run. This design is 1.20\% below the best TPE seed and $\sim$2\% below the TPE median; with respect to the $(1+1)$-ES, it is 0.23\% below the best seed and 0.9\% below the median. It is obtained with at most 50 simulator calls per run.
The median final value of Haiku~5.5 is only 0.012\% above $\mathrm{NPC}^{\star}$, and is still below the medians of TPE and of the $(1+1)$-ES after 300 evaluations; the feasible rate of all three agents is higher than that of any baseline.
Opus~5.5 ends at the best design in every run, Sonnet~5.5 in most runs and Haiku~5.5 only once. In terms of near-optimal designs, however, Haiku is in these runs more reliable than Sonnet: all its runs reach the $1\%$ band, whereas two Sonnet runs never do, because they remain trapped with a larger heat pump. Haiku is also the slowest of the three models to reach the band.
Compared with the only baseline that reaches the band, the $(1+1)$-ES, the agent reaches it more often and, among the runs that reach it, needs between seven and eleven times fewer evaluations.
The gap between the best agent design and the best TPE seed is about 43~kEUR over 25 years, and that to the best $(1+1)$-ES seed about 8~kEUR; the gaps to the medians are about 71~kEUR (TPE) and 32~kEUR ($(1+1)$-ES). These differences are small compared with the uncertainty introduced by the modelling assumptions discussed in Section~\ref{sec:case_study}. The result is not that the agent finds a materially better plant than a well-run Bayesian optimizer or local search, but that it reaches the best region of the design space with a few tens of simulations in almost all runs, whereas the best baseline reaches it in half of its seeds and only after about two hundred evaluations, while producing a motivation for each decision.

\begin{table}[pos=htbp]
\centering
\caption{Best design found by each method over its runs (seeds). The three agents reach the same design and NPC; their PV inverter is 360 or 400~kW, which gives the same NPC. Efficiencies in parentheses; the CHP of the agents has $\eta_\mathrm{el}=0.32$ and $\eta_\mathrm{th}=0.58$; -- means not installed.}
\label{tab:best_solutions}
\footnotesize
\setlength{\tabcolsep}{4pt}
\begin{tabular}{@{}lcccc@{}}
\toprule
 & $(1+1)$-ES & \hd{Random\\search} & TPE & \hd{Agents\\(all models)} \\
\midrule
NPC [EUR] & 3,533,721 & 3,756,201 & 3,568,489 & 3,525,522 \\
\midrule
PV field & 2,000\,m$^2$ & 1,690\,m$^2$ & 2,000\,m$^2$ & 2,000\,m$^2$ \\
PV inverter & 341\,kW (0.98) & 245\,kW (0.98) & 293\,kW (0.98) & 360--400\,kW (0.98) \\
Wind turbine & -- & -- & -- & -- \\
CHP & -- & -- & -- & 10\,kW$_\mathrm{el}$ \\
Solar thermal & 265\,m$^2$ & 345\,m$^2$ & 125\,m$^2$ & 260\,m$^2$ \\
Boiler & -- & -- & -- & -- \\
Battery capacity & 100\,kWh & 90\,kWh & 90\,kWh & 100\,kWh \\
Battery power & 22.5\,kW (0.99) & 47.5\,kW (0.84) & 30\,kW (0.98) & 50\,kW (0.99) \\
Battery inverter & 35\,kW (0.96) & 50\,kW (0.98) & 11\,kW (0.98) & 50\,kW (0.98) \\
Heat pump & 700\,kW$_\mathrm{th}$ & 700\,kW$_\mathrm{th}$ & 700\,kW$_\mathrm{th}$ & 700\,kW$_\mathrm{th}$ \\
\bottomrule
\end{tabular}
\end{table}

\subsection{Resulting designs} 
Table \ref{tab:best_solutions} presents the best design found by each optimisation method.
The designs are qualitatively consistent across methods. The best design, found by all ten Opus~5.5 runs, six Sonnet~5.5 runs and one Haiku~5.5 run, combines a reversible heat pump of 700~kW\textsubscript{th}, a PV field of 2{,}000~m$^2$ with an inverter of 360 or 400~kW (the two ratings give the same NPC), a battery of 100~kWh and 50~kW, a solar-thermal field of 260~m$^2$ and a CHP unit of 10~kW\textsubscript{el}; no boiler and no wind turbine are installed.
Two aspects of the best design deserve comment. First, the PV area is at the upper bound of its admissible range, and so are the battery capacity and power; these bounds are effectively active, and a larger admissible range, or an explicit roof-area constraint, would change the result. The PV inverter has no cost in this scenario, and every size above about 340~kW gives the same NPC, so the optimum is degenerate in this variable: what matters is that the inverter can handle the PV peak at the maximum irradiance, $2{,}000~\mathrm{m^2}\times0.18\times937~\mathrm{W/m^2}\approx337$~kW.
Second, all conversion efficiencies are at their upper bounds, as expected from a cost model that does not link cost to efficiency directly. The battery is part of all the best designs, and it is valued only through self-consumption, since the tariff is flat and exported energy is paid 0.08~EUR/kWh against 0.30~EUR/kWh for purchased energy.
The other agent runs end on close variants of this design. Five Haiku runs omit the CHP and use 270~m$^2$ of solar thermal (+425~EUR); the remaining near-optimal runs differ only in the solar-thermal area (250--255~m$^2$) or in the CHP size (30~kW\textsubscript{el} in one Sonnet run); the Haiku run that stopped early also keeps a 5~kW boiler (+11.7~kEUR), and the worst Haiku run has no CHP, 300~m$^2$ of solar thermal and a smaller battery (70~kWh, 30~kW; +18.2~kEUR). The two Sonnet runs that do not reach the 1\% band have a similar design except for the heat pump, which is 800~kW (+136~kEUR). Every agent run installs the 2{,}000~m$^2$ PV field. 
All ten TPE seeds and all ten $(1+1)$-ES seeds also select a 700~kW heat pump without boiler, together with solar thermal; the PV field is 1{,}950--2{,}000~m$^2$ in all TPE seeds and in nine $(1+1)$-ES seeds (560~m$^2$ in the remaining one). The battery is selected in seven TPE and nine $(1+1)$-ES seeds, the CHP in five TPE seeds and in none of the $(1+1)$-ES seeds, and a small wind turbine (1.5--4.5~kW) in three TPE seeds. Random search finds the 700~kW heat pump in only two seeds out of ten and otherwise settles on larger machines of 750--950~kW.

\subsection{Ablation: Removing the Domain-Knowledge File}
\label{subsec:ablation_insights}

\begin{figure}[pos=htbp]
\centering
\begin{minipage}[t]{0.49\textwidth}
\centering
\includegraphics[width=\textwidth]{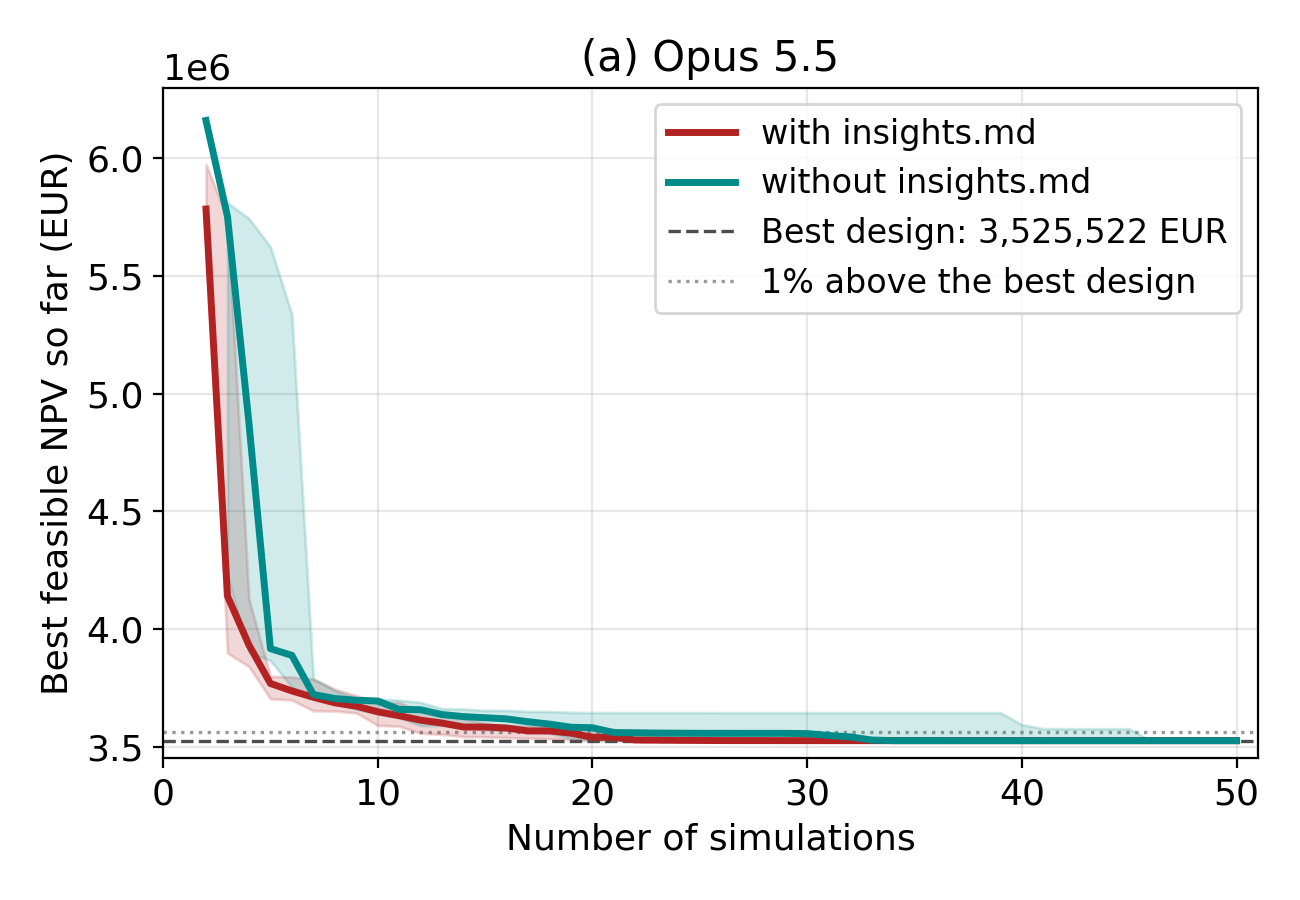}\\
{\small (a)}
\end{minipage}\hfill
\begin{minipage}[t]{0.49\textwidth}
\centering
\includegraphics[width=\textwidth]{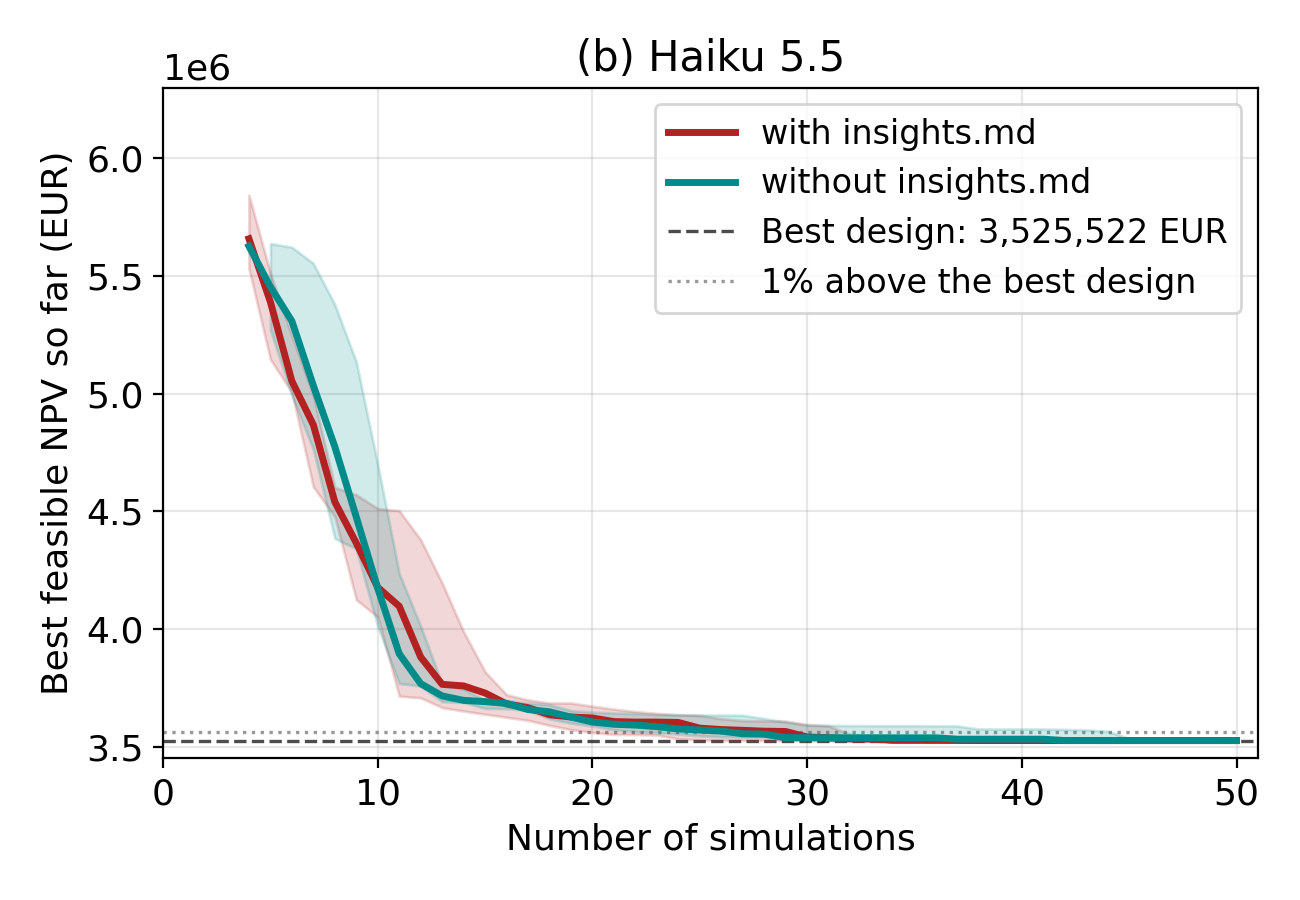}\\
{\small (b)}
\end{minipage}
\caption{Best feasible NPC found within the first $x$ evaluations by (a) Opus~5.5 and (b) Haiku~5.5 with and without the domain-knowledge file \texttt{insights.md}. Median and interquartile range over ten runs, computed as in Figure~\ref{fig:comparison_results}. The dashed line is $\mathrm{NPC}^{\star}$ and the dotted line is $1.01\,\mathrm{NPC}^{\star}$.}
\label{fig:ablation_insights}
\end{figure}

To measure the contribution of \texttt{insights.md} (Section~\ref{subsubsec:instructions}), ten Opus~5.5 runs and ten Haiku~5.5 runs were repeated without access to it, with the same initial configuration, budget, hooks and acceptance rule. None of the ablation sessions reads the file, as verified on the session transcripts.
Figure~\ref{fig:ablation_insights} compares the convergence curves of the two configurations for each model, and Table~\ref{tab:ablation_insights} summarises the results.

\begin{table}[pos=htbp]
\centering
\caption{Opus~5.5 and Haiku~5.5 with and without the domain-knowledge file \texttt{insights.md}: final NPC (EUR) over ten runs of 50 simulations, feasible rate, evaluations to come within 1\% of $\mathrm{NPC}^{\star}$ (as in Table~\ref{tab:multi_seed}), and runs ending at the best design.}
\label{tab:ablation_insights}
\footnotesize
\setlength{\tabcolsep}{4pt}
\begin{tabular}{@{}lccccc@{}}
\toprule
Configuration & Best & Median & \hd{Feasible\\rate} & \hd{Evals to 1\%\\of best} & \hd{Runs\\at best} \\
\midrule
\multicolumn{6}{@{}l}{\emph{Opus~5.5}} \\
With \texttt{insights.md} & 3,525,522 & 3,525,522 & 89.0\% & 10/10 $\cdot$ 18 (31) & 10/10 \\
Without \texttt{insights.md} & 3,525,522 & 3,525,522 & 90.0\% & 8/10 $\cdot$ 19.5 (46) & 8/10 \\
\midrule
\multicolumn{6}{@{}l}{\emph{Haiku~5.5}} \\
With \texttt{insights.md} & 3,525,522 & 3,525,947 & 87.6\% & 10/10 $\cdot$ 27.5 (45) & 1/10 \\
Without \texttt{insights.md} & 3,525,522 & 3,525,947 & 89.0\% & 9/10 $\cdot$ 27 (50) & 2/10 \\
\bottomrule
\end{tabular}
\end{table}

Without the file, Opus~5.5 still finds $\mathrm{NPC}^{\star}$, and the median run ends on it; the feasible rate is essentially unchanged. Eight runs end at the best design, and the number of evaluations needed to enter the 1\% band is similar to that of the complete configuration (median 19.5 against 18, maximum 46 against 31; Figure~\ref{fig:ablation_insights}a). Without the file the first experiments are also less effective: after five simulations the median best NPC is 3.92~MEUR, against 3.77~MEUR with the file, and the two curves become similar only after about seven simulations.
Also for Haiku~5.5 the differences are small (Figure~\ref{fig:ablation_insights}b). The final values are the same as with the file (best $\mathrm{NPC}^{\star}$, median 3{,}525{,}947~EUR), two runs end at the best design instead of one, nine of ten enter the 1\% band after a median of 27 evaluations (27.5 with the file), and the two convergence curves are close from the first experiments.
The domain-knowledge file is therefore not needed to find the best design: both models reach $\mathrm{NPC}^{\star}$ and the same median value without it, inferring from the simulator output, which prints the CAPEX and OPEX at every run, most of what the file states, including the absence of costs for efficiencies and inverters. The file appears, however, to reduce the variability between runs (Figure~\ref{fig:ablation_insights}). For Opus~5.5, the interquartile range of the best NPC is narrower with the file in the first experiments and closes after about 30 simulations, whereas without the file it remains at about 120~kEUR until simulation~35. For Haiku~5.5 the effect is weaker and not uniform: the range is narrower with the file between the sixth and the tenth simulation and after the 35th, but wider between the 12th and the 20th. These results suggest a more predictable search, especially for Opus~5.5.

\section{Discussion}
\label{sec:discussion}

In this case study the agent reaches designs of equal or better final quality than the baselines with far fewer simulations (Table~\ref{tab:multi_seed}).
The $(1+1)$-ES shares the agent's starting configuration, acceptance rule and limit of one to three changes per move (Section~\ref{sec:experiments}), and differs only in how the move is chosen: the ES perturbs randomly chosen parameters independently, with random step sizes and random sizes for a technology switched on, whereas the agent chooses related parameters together, such as a PV field and its inverter, or a boiler and the heat-pump size. Since the acceptance rule is the same, the difference in the number of evaluations is due to the choice of moves, and this is precisely where the agent differs from a black-box optimizer: it exploits information that the baselines cannot use, namely the cost breakdown returned by the simulator, the domain knowledge of \texttt{insights.md} and the general design knowledge of the model, and turns it into coordinated moves.

The model affects the reliability and the speed of the search: Opus~5.5 reaches the best design in every run, whereas two Sonnet~5.5 runs never come within 1\% of it. These two runs are held by a conclusion drawn early, in a design without other heat sources, and never retested (Appendix~\ref{app:qualitative}): since the acceptance rule never returns to a worse design, such a conclusion is corrected only if the agent itself decides to test it. 
Every configuration, outcome and decision of the agent is logged and can be reproduced. The explanations recorded in the notes, however, are not always correct (Appendix~\ref{app:qualitative}).
The approach reduces the number of simulator calls, not the total computation: every experiment also requires a call to the language model, which for cloud-served models has a cost. A run cost between 1.61 and 2.26~USD with Opus~5.5, between 0.99 and 1.75~USD with Sonnet~5.5 and between 0.63 and 1.28~USD with Haiku~5.5, and lasted between 11 and 17~minutes, except for two Sonnet sessions (37~minutes and 2.4~hours) in which the agent inserted unnecessary waits. In this case study one simulation takes about 7~s and a baseline seed of 300 evaluations about 35~minutes, so in monetary terms the agent is not cheaper than the baselines, although a run takes less wall-clock time than a baseline seed or than the about 22~minutes that the $(1+1)$-ES needs, in its median successful seed, to enter the 1\% band; its smaller number of simulations reduces the total cost only when a single simulation is considerably more expensive than a call to the model.
The results have the following limitations. They refer to a single case study, with one climate, one tariff structure and a single-objective NPC formulation. They are obtained with one family of models and one coding agent, whose versions are reported in Section~\ref{sec:experiments}; other models and agents were not tested. 
The agent receives in \texttt{insights.md} the cost structure of the technologies, from which it can infer, for instance, that efficiencies and inverter ratings carry no cost, information that the black-box baselines do not use; since the cost model does not depend on the efficiencies, their optimum lies at the upper bound. 
The ablation of Section~\ref{subsec:ablation_insights} shows, however, that this information is not needed to find the best design, since the agent can infer it from the simulator output.
Finally, part of the agent's behaviour may reflect design rules learned during training rather than reasoning on the observed results, and the present experiments cannot separate the two.
Natural extensions are an acceptance rule that tolerates temporary worsening of NPC to allow more exploration; multi-objective formulations; the evaluation of other models (e.g., local models with reasoning capabilities) and other harnessings.

\section{Conclusions}
\label{sec:conclusions}

This paper presented AutoEnergyDesign, a framework for automatic energy system design optimization driven by Large Language Models. The main contributions are an LLM-based optimization architecture organised around an explicit set of memories, a detailed techno-economic formulation of the design problem, a git-based state-management mechanism that never loses the best design found, and an empirical comparison against established baselines with three models of different capability.
Overall, the results indicate that LLM-driven optimization can reach competitive solution quality at a substantially reduced number of simulator evaluations while offering a level of traceability of the decisions that is difficult to obtain from conventional black-box methods alone.
In the case study, the agents driven by Haiku~5.5, Sonnet~5.5 and Opus~5.5 all found the best design of the study, with an NPC of about 3.526~MEUR. This value is 1.20\% below the best of ten TPE seeds and 0.23\% below the best of ten $(1+1)$-ES seeds of 300 evaluations, and was obtained with at most 50 simulations per run. 
The models differ in reliability and speed: Opus~5.5 reached the best design in all ten runs, coming within 1\% of it after a median of 18 simulations; Sonnet~5.5 reached it in six runs, while two never came within 1\% of it; Haiku~5.5 reached it exactly only once, but all its runs came within 1\% of it, after a median of 27.5 simulations. 
Without the domain-knowledge file, Opus~5.5 still found the best design, in eight of ten runs: the file is not needed to reach it, but it seems to make the search more predictable, especially for Opus~5.5. For Haiku~5.5, whose limit is the final refinement of the design, removing the file makes practically no difference.
Among the baselines, only the $(1+1)$ evolution strategy, which shares the agent's acceptance rule, comes within 1\% of the best design, in five of ten seeds and after a median of 192 evaluations, which suggests that the advantage comes from the moves chosen by the model rather than from the acceptance rule: the agent exploits information on the problem that a black-box optimizer cannot use. The reasoning notes make every decision auditable and, in this sense, AutoEnergyDesign should be considered not as a replacement for existing optimizers, but as a complementary paradigm for engineering design problems in which reasoning traces, domain knowledge, and iterative experimentation all play a central role.

\printcredits

\section*{Data availability}
The data are not publicly available due to privacy restrictions. The data may be available on request from the corresponding author, subject to evaluation and approval.
\section*{Declaration of generative AI and AI-assisted technologies in the manuscript preparation process}
During the preparation of this work the authors used Claude in order to improve the grammar and phrasing of the manuscript. After using this tool/service, the authors reviewed and edited the content as needed and take full responsibility for the content of the published article.

\bibliographystyle{elsarticle-num}

\bibliography{references}

%
%

\newpage
\def\inmain{}
\section*{Appendices}
\ifx\inmain\undefined
\documentclass[a4paper,fleqn,review]{cas-sc}

\usepackage[utf8]{inputenc}
\usepackage{longtable}
\usepackage{float}
\usepackage{enumitem}
\setlist{nosep}

\usepackage[numbers,sort&compress]{natbib}

\geometry{paperwidth=210mm, paperheight=297mm, top=43mm, bottom=43mm, left=48mm, right=48mm}

\setlength{\emergencystretch}{3em}
\hbadness=10000

\ExplSyntaxOn
\cs_set:Npn \__make_tbl_caption:nn #1#2
{
  \l_tbl_align_tl
  \skip_vertical:N \l_tbl_abovecap_skip
  {\parbox{ \dimexpr(\l_tbl_width_dim)}
    {\rightskip=0pt\sffamily\footnotesize\textbf{\color{scolor}#1}\par#2\par\vskip4pt }}
  \skip_vertical:N \l_tbl_belowcap_skip
}
\cs_set:Npn \__make_fig_caption:nn #1#2
{
  \l_fig_align_tl
  \skip_vertical:N \l_fig_abovecap_skip
  \setbox\cascaptionbox=\hbox{%
     \sffamily\footnotesize\textbf{\color{scolor}#1:}~#2}
  \ifdim\the\wd\cascaptionbox<\dim_use:N \l_fig_width_dim\relax
    \parbox{ \l_fig_width_dim }
      {\unskip\ignorespaces\hfil\sffamily\footnotesize
       \textbf{\color{scolor}#1:}~#2\hfil\par }
  \else
    \parbox{ \l_fig_width_dim }
      {\rightskip=0pt\unskip\ignorespaces\sffamily
       \footnotesize\textbf{\color{scolor}#1:}~#2\par }
  \fi
  \skip_vertical:N \l_fig_belowcap_skip
}
\makeatletter
\RenewDocumentEnvironment { table } { O{} }
{
  \__reset_tbl:
  \bool_gset_false:N \g_tbl_full_bool
  \dim_set:Nn \l_tbl_width_dim { \linewidth }
  \keys_set:nn { cas / tbl } { #1 }
  \csxdef{fps@table}{\l_tbl_pos_tl}
  \csgdef{tblwidth}{\dim_use:N \l_tbl_width_dim}
  \cs_set_eq:NN \@makecaption \__make_tbl_caption:nn
  \@float{table}
  \l_tbl_align_tl
  \sffamily\normalsize
}
{ \end@float }
\RenewDocumentEnvironment { table* } { O{width=\FullWidth} }
{
  \__reset_tbl:
  \bool_gset_true:N \g_tbl_full_bool
  \dim_set:Nn \l_tbl_width_dim { \FullWidth }
  \keys_set:nn { cas / tbl } { #1 }
  \csxdef{fps@table}{\l_tbl_pos_tl}
  \csgdef{tblwidth}{\dim_use:N \l_tbl_width_dim}
  \cs_set_eq:NN \@makecaption \__make_tbl_caption:nn
  \@dblfloat{table}
  \l_tbl_align_tl
  \sffamily\normalsize
}
{ \end@dblfloat }
\makeatother
\ExplSyntaxOff

\usepackage{xr}
\externaldocument[M-]{Main}
\newcommand{\mref}[1]{\ref{M-#1}}  

\begin{document}
\let\WriteBookmarks\relax
\def\floatpagepagefraction{1}
\def\textpagefraction{.001}
\shorttitle{AutoEnergyDesign}
\shortauthors{A. Buonanno et~al.}

\begin{center}
{\LARGE Appendices to ``AutoEnergyDesign: An LLM-Driven Agent for Automatic Energy System Design Optimization''\par}
\vspace{1em}
{\large Amedeo Buonanno, Salvatore Fabozzi, Maria Valenti\par}
\end{center}
\doublespacing
\fi
\providecommand{\mref}[1]{\ref{#1}}

\appendix
\counterwithin{equation}{section}
\counterwithin{table}{section}
\counterwithin{figure}{section}
\section{Nomenclature}\label{app:nomen}

Table~\ref{tab:nomenclature} summarizes the variables and symbols used throughout the mathematical formulation.

\begingroup
\hfuzz=500pt
\hbadness=10000
\small
\setlength{\LTleft}{0pt}
\setlength{\LTright}{0pt}
\begin{longtable}{@{}>{\raggedright\arraybackslash}p{0.22\textwidth} >{\raggedright\arraybackslash}p{0.44\textwidth} >{\raggedright\arraybackslash}p{0.13\textwidth}@{}}
\caption[Nomenclature]{Variables and symbols used in the model}\label{tab:nomenclature} \\
\toprule
\textbf{Symbol} & \textbf{Description} & \textbf{Unit} \\
\midrule
\endfirsthead
\toprule
\textbf{Symbol} & \textbf{Description} & \textbf{Unit} \\
\midrule
\endhead
\bottomrule
\endfoot
$\mathbf{x}$ & Design decision vector (technology activation and sizing) & -- \\
$\mathbf{u}_{1:T}$ & Sequence of operational decision vectors over the horizon & -- \\
$t$ & Time index (hour) & h \\
$T$ & Total number of simulation time steps & h \\
$d_t^{\mathrm{el}}$ & Exogenous electricity demand at time $t$ & kW \\
$d_t^{\mathrm{heat}}$ & Exogenous heat demand (space heating + DHW) at time $t$ & kW$_{\mathrm{th}}$ \\
$d_t^{\mathrm{cool}}$ & Exogenous cooling demand at time $t$ & kW$_{\mathrm{th}}$ \\
$p_t^{\mathrm{sup}}$ & Electricity supplied to demand at time $t$ & kW \\
$q_t^{\mathrm{heat}}$ & Heat supplied to demand at time $t$ & kW$_{\mathrm{th}}$ \\
$q_t^{\mathrm{cool}}$ & Cooling power supplied to demand at time $t$ & kW$_{\mathrm{th}}$ \\
$s_t^{\mathrm{el}}, s_t^{\mathrm{heat}}, s_t^{\mathrm{cool}}$ & Unserved electricity, heat and cooling at time $t$ (also written $s^{\text{el}}(t)$, etc.) & kW, kW$_{\mathrm{th}}$ \\
$\mathcal{T}$ & Set of available technologies & -- \\
$\mathcal{X}$ & Feasible set of design decisions & -- \\
$g(\mathbf{x},\mathbf{u}_{1:T})$ & Vector of technical and operational constraints & -- \\
$z_{\mathrm{pv}}$ & Binary activation variable for PV technology & -- \\
$z_{\mathrm{wt}}$ & Binary activation variable for wind-turbine technology & -- \\
$z_{\mathrm{chp}}$ & Binary activation variable for CHP technology & -- \\
$z_{\mathrm{st}}$ & Binary activation variable for solar-thermal technology & -- \\
$z_{\mathrm{boi}}$ & Binary activation variable for boiler technology & -- \\
$z_{\mathrm{bat}}$ & Binary activation variable for battery technology & -- \\
$z_{\mathrm{hp}}$ & Binary activation variable for heat-pump technology & -- \\
$A_{\mathrm{pv}}$ & Installed photovoltaic collector area & m$^2$ \\
$P_{\mathrm{wt}}^{\mathrm{rated}}$ & Rated wind-turbine electrical power & kW \\
$P_{\mathrm{chp}}^{\mathrm{el}}$ & CHP nominal electrical power & kW \\
$\eta_{\mathrm{chp}}^{\mathrm{el}}$ & CHP electrical efficiency & -- \\
$\eta_{\mathrm{chp}}^{\mathrm{th}}$ & CHP thermal efficiency & -- \\
$A_{\mathrm{st}}$ & Installed solar-thermal collector area & m$^2$ \\
$P_{\mathrm{boi}}^{\max}$ & Maximum boiler thermal power & kW$_{\mathrm{th}}$ \\
$\eta_{\mathrm{boi}}$ & Boiler thermal efficiency & -- \\
$E_{\mathrm{bat}}^{\max}$ & Battery energy capacity & kWh \\
$P_{\mathrm{bat}}^{\max}$ & Battery maximum charge/discharge power & kW \\
$\eta_{\mathrm{bat}}$ & Battery round-trip efficiency & -- \\
$P_{\mathrm{inv},j}^{\max}$ & Nominal power of inverter $j\in\{\mathrm{pv},\mathrm{bat},\mathrm{wt}\}$ & kW \\
$\eta_{\mathrm{inv},j}$ & Conversion efficiency of inverter $j$ & -- \\
$P_{\mathrm{hp}}^{\max}$ & Heat-pump maximum thermal power & kW$_{\mathrm{th}}$ \\
$\mathrm{EER}$ & Heat-pump cooling energy-efficiency ratio (reversible machine) & -- \\
$\mathrm{GHI}(t)$ & Global horizontal irradiance at time $t$ & W/m$^2$ \\
$p_{\text{pv}}(t)$ & PV electrical output at time $t$ & kW \\
$\eta_{\text{pv}}$ & PV conversion efficiency & -- \\
$v,\ v(t)$ & Wind speed (generic; at time $t$) & m/s \\
$v_{\text{in}}, v_{\text{rated}}, v_{\text{out}}$ & Wind turbine cut-in, rated, and cut-out speeds & m/s \\
$P_{\text{wind}}(v)$ & Wind-turbine electrical output as a function of $v$ & kW \\
$p_{\text{wind}}(t)$ & Wind-turbine electrical output at time $t$ & kW \\
$p_{\text{chp}}^{\text{el}}(t),\ p_{\text{chp}}^{\text{th}}(t)$ & CHP electrical and thermal output at time $t$ & kW, kW$_{\mathrm{th}}$ \\
$\sigma$ & CHP heat-to-power ratio, $\sigma=\eta_{\mathrm{chp}}^{\mathrm{th}}/\eta_{\mathrm{chp}}^{\mathrm{el}}$ & -- \\
$\hat p_{\text{st}}(t)$ & Available solar-thermal heat output at time $t$ & kW$_{\mathrm{th}}$ \\
$p_{\text{st}}(t)$ & Solar-thermal heat actually used at time $t$ & kW$_{\mathrm{th}}$ \\
$\eta_{\text{st}}$ & Solar-thermal collector efficiency & -- \\
$p_{\text{boiler}}^{\text{gas}}(t)$ & Boiler gas consumption at time $t$ & kW$_{\mathrm{fuel}}$ \\
$c_{\text{boiler}}$ & Boiler heat marginal cost & EUR/kWh$_{\mathrm{th}}$ \\
$c_{\text{gas}}$ & Natural gas price & EUR/kWh \\
$\eta_{\text{store}}, \eta_{\text{dispatch}}$ & Battery charging/discharging efficiencies & -- \\
$p_{\text{batt}}^{\text{charge}}(t)$, $p_{\text{batt}}^{\text{discharge}}(t)$ & Battery charging/discharging powers at time $t$ & kW \\
$\mathrm{COP}$ & Heat-pump coefficient of performance (fixed, heating) & -- \\
$p_{\text{hp}}^{\text{th}}(t)$ & Heat-pump heat output at time $t$ & kW$_{\mathrm{th}}$ \\
$p_{\text{hp,c}}^{\text{el}}(t),\ p_{\text{hp,c}}^{\text{cool}}(t)$ & Heat-pump electrical input and cooling output in cooling mode & kW, kW$_{\mathrm{th}}$ \\
$c_{\text{hp}}$ & Heat-pump heat marginal cost & EUR/kWh$_{\mathrm{th}}$ \\
$c_{\text{elec}}$ & Electricity purchase price & EUR/kWh \\
$p^{\text{in}}_{\mathrm{inv},j}(t),\ p^{\text{out}}_{\mathrm{inv},j}(t)$ & Source-side input and delivered output power of inverter $j$ & kW \\
$c_{\text{chp}}^{\text{var}}$ & CHP variable maintenance cost & EUR/kWh$_{\mathrm{el}}$ \\
$y$ & Year index in the economic formulation & yr \\
$\mathrm{PV}(C,y)$ & Present value of a cash flow $C$ occurring at year $y$ & EUR \\
$r$ & Annual discount rate & -- \\
$N$ & Project horizon in years & yr \\
$\mathrm{PV}_{\text{factor}}$ & Annuity present-value factor over the horizon & -- \\
$\mathrm{CAPEX}_i$ & Initial capital expenditure of technology $i$ & EUR \\
$\mathcal{R}_i$ & Set of replacement years for technology $i$ & yr \\
$\mathrm{CAPEX}_{\text{total}}$ & Total discounted CAPEX including replacements & EUR \\
$f_{\text{repl}}(L,N,r)$ & Replacement multiplier including discounting & -- \\
$L, L_i$ & Lifetime (generic or of technology $i$) & yr \\
$m$ & Replacement index in discounted replacement sum & -- \\
$\mathrm{CAPEX}_{\text{effective},i}$ & Effective discounted CAPEX for technology $i$ over the horizon & EUR \\
$\mathrm{OPEX}_{\text{fixed},i}$ & Annual fixed OPEX for technology $i$ & EUR/yr \\
$\mathrm{OPEX}_{\text{fixed}}^{\text{PV}}$ & Present value of fixed OPEX over the horizon & EUR \\
$\mathrm{OPEX}_{\text{variable}}$ & Annual variable operating cost from dispatch & EUR/yr \\
$\mathrm{NPC}$ & Total discounted life-cycle cost (costs net of revenue) & EUR \\
$p_{\text{grid}}^{\text{import}}(t)$ & Electricity imported from the grid at time $t$ & kW \\
$p_{\text{grid}}^{\text{export}}(t)$ & Electricity exported to the grid at time $t$ & kW \\
$p_{\text{boiler}}(t)$ & Boiler thermal output at time $t$ & kW$_{\mathrm{th}}$ \\
$p_{\text{chp}}^{\text{gas}}(t)$ & CHP gas consumption at time $t$ & kW$_{\mathrm{fuel}}$ \\
$p_{\text{hp}}^{\text{el}}(t)$ & Heat-pump electrical input in heating mode at time $t$ & kW \\
$\mathrm{soc}(t)$ & Battery state of charge at time $t$ & kWh \\
$p_{\text{load}}^{\text{el}}(t)$ & Electrical load in the dispatch & kW \\
$p_{\text{load}}^{\text{heat}}(t), p_{\text{load}}^{\text{dhw}}(t)$ & Space-heating and DHW loads in the dispatch & kW$_{\mathrm{th}}$ \\
$p_{\text{load}}^{\text{cool}}(t)$ & Cooling load in the dispatch & kW$_{\mathrm{th}}$ \\
$p(t),\ P^{\text{rated}}$ & Generic dispatch variable of a technology and its rated limit & kW \\
$c_{\text{export}}$ & Electricity export tariff in dispatch objective & EUR/kWh \\
$\pi$ & Penalty cost for unserved energy & EUR/kWh \\
$\varepsilon$ & Tolerance on annual unserved energy for feasibility & kWh \\
$E_{\mathrm{grid}}^{\mathrm{import}}, E_{\mathrm{el}}^{\mathrm{tot}}$ & Annual grid import and total electrical consumption & kWh \\
$S_i$ & Discounted salvage (residual) value of technology $i$ at year $N$ & EUR \\
$y_i^{\text{last}}$ & Year of the last installation of technology $i$ & yr \\
$B$ & Evaluation budget (number of simulator calls) & -- \\
$N_a$ & Number of experiments between two reflective synthesis turns & -- \\
$K$ & Number of most recent discarded designs listed in the search history & -- \\
$k$ & Experiment index & -- \\
$\mathbf{x}_k,\ \mathbf{x}^\star_k$ & Design proposed at experiment $k$; best valid design after experiment $k$ & -- \\
$\sigma_{\mathrm{ES}},\ \sigma_{\mathrm{ES},0}$ & Mutation step of the $(1+1)$-ES as a fraction of the grid levels; initial value & -- \\
$\mathrm{NPC}^{\star}$ & Best NPC found in the study (by any method and run) & EUR \\
$\delta$ & Relative tolerance on $\mathrm{NPC}^{\star}$ used to count the evaluations to a near-optimal design & -- \\
\end{longtable}
\endgroup

\section{Technology models}\label{app:tech}

\paragraph{Photovoltaic Solar}

PV generation is modeled as:
\begin{equation}
    p_{\text{pv}}(t) = \text{GHI}(t) \times A_{\text{pv}} \times \eta_{\text{pv}} \times \frac{1}{1000}
\end{equation}
where $\text{GHI}(t)$ is solar irradiance, $A_{\text{pv}}$ is installed area, and $\eta_{\text{pv}}$ is module efficiency, fixed at $0.18$. Investment cost scales linearly with area, at $121.3$~EUR/m$^2$ (Table~\ref{tab:capex}).

\paragraph{Micro-Wind Turbine}

Wind generation follows a standard piecewise power curve:

\begin{equation}
P_{\text{wind}}(v) =
\begin{cases}
0 & v < v_{\text{in}} \\
P_{\mathrm{wt}}^{\mathrm{rated}} \times \left(\frac{v - v_{\text{in}}}{v_{\text{rated}} - v_{\text{in}}}\right)^3 & v_{\text{in}} \leq v < v_{\text{rated}} \\
P_{\mathrm{wt}}^{\mathrm{rated}} & v_{\text{rated}} \leq v \leq v_{\text{out}} \\
0 & v > v_{\text{out}}
\end{cases}
\end{equation}
where $v_{\text{in}}$, $v_{\text{rated}}$, and $v_{\text{out}}$ denote cut-in, rated, and cut-out speed, respectively. The hourly output entering the dispatch is $p_{\text{wind}}(t)=P_{\text{wind}}(v(t))$, with $v(t)$ the wind speed at hour $t$. The available installation area constrains any prospective wind turbines to a fixed hub height of 10 m, which also corresponds to the reference height of the available wind-speed measurements. This does not represent a significant limitation, since standard vertical extrapolation laws can be applied to estimate wind speeds at different heights from data measured at 10 m.

\paragraph{Combined Heat and Power (CHP)}

CHP units co-generate electricity and heat from gas input:

\begin{align}
    p_{\text{chp}}^{\text{el}}(t) &= \eta_{\mathrm{chp}}^{\mathrm{el}}\, p_{\text{chp}}^{\text{gas}}(t) \\
    p_{\text{chp}}^{\text{th}}(t) &= \eta_{\mathrm{chp}}^{\mathrm{th}}\, p_{\text{chp}}^{\text{gas}}(t)
\end{align}
where $p_{\text{chp}}^{\text{gas}}(t)$ is the gas input power and $\eta_{\mathrm{chp}}^{\mathrm{el}}$ and $\eta_{\mathrm{chp}}^{\mathrm{th}}$ are the electrical and thermal efficiencies. This directly couples electric and thermal outputs, with heat-to-power ratio $\sigma=\eta_{\mathrm{chp}}^{\mathrm{th}}/\eta_{\mathrm{chp}}^{\mathrm{el}}$.

Although the two efficiencies are independent design variables, their admissible ranges (Table~\ref{tab:design_bounds}) bound the total CHP efficiency to at most $0.90$, thereby excluding physically impossible configurations with $\eta_{\mathrm{chp}}^{\mathrm{el}}+\eta_{\mathrm{chp}}^{\mathrm{th}}>1$.

\paragraph{Solar Thermal}

Solar thermal output is modeled as:
\begin{equation}
    \hat p_{\text{st}}(t) = \text{GHI}(t) \times A_{\text{st}} \times \eta_{\text{st}} \times \frac{1}{1000}
\end{equation}
where $A_{\text{st}}$ and  $\eta_{\text{st}}$ are the collector area and its efficiency, respectively. The heat actually used, $p_{\text{st}}(t)\le\hat p_{\text{st}}(t)$, is a dispatch variable (Section~\mref{sec:dispatch}).

\paragraph{Gas Boiler}

The boiler converts gas to useful heat as:
\begin{equation}
    p_{\text{boiler}}(t) = \eta_{\mathrm{boi}}\, p_{\text{boiler}}^{\text{gas}}(t)
\end{equation}
and its marginal heat cost is:
\begin{equation}
    c_{\text{boiler}} = \frac{c_{\text{gas}}}{\eta_{\mathrm{boi}}}
\end{equation}
where $p_{\text{boiler}}^{\text{gas}}(t)$ is the gas input power and $c_{\text{gas}}$ is the gas price.

\paragraph{Battery Storage}

Battery storage is represented through charge, discharge, and state-of-charge dynamics. The round-trip efficiency $\eta_{\text{bat}}$ is split symmetrically between charging and discharging:
\begin{align}
    \eta_{\text{store}} &= \sqrt{\eta_{\text{bat}}} \\
    \eta_{\text{dispatch}} &= \sqrt{\eta_{\text{bat}}}
\end{align}

The state of charge evolves as:
\begin{equation}
    \mathrm{soc}(t+1) = \mathrm{soc}(t) + \eta_{\text{store}}\, p_{\text{batt}}^{\text{charge}}(t) - \frac{p_{\text{batt}}^{\text{discharge}}(t)}{\eta_{\text{dispatch}}}
\end{equation}
subject to power and energy capacity limits.

\paragraph{Heat Pump}

Heat-pump performance is expressed through the coefficient of performance (COP):
\begin{equation}
    \text{COP} = \frac{p_{\text{hp}}^{\text{th}}(t)}{p_{\text{hp}}^{\text{el}}(t)}
\end{equation}

The corresponding marginal heat cost is:
\begin{equation}
    c_{\text{hp}} = \frac{c_{\text{elec}}}{\text{COP}}
\end{equation}

where $c_{\text{elec}}$ is the electricity price. In the present implementation the COP is a fixed parameter (COP $=3.5$) and does not vary with operating conditions. The same machine is \emph{reversible}: when cooling is required, it delivers cooling instead of heat, modelled as a second electricity-to-cooling conversion with a fixed energy-efficiency ratio (EER $=3.0$). The rated thermal power $P_{\text{hp}}^{\max}$ applies to both modes, and the cooling capability incurs no additional CAPEX or fixed OPEX. A single compressor, however, serves both modes: within each hour the machine can only share its time between heating and cooling, so that the heat and the cooling it delivers in the same hour cannot together exceed $P_{\text{hp}}^{\max}$ (Eq.~\eqref{eq:hp_timeshare}). In the case study, space heating and space cooling never occur in the same hour, but domestic hot water, which is part of the heat demand, can coincide with cooling. In those hours the heat pump must be sized for the combined demand, unless another heat source supplies part of the hot water; since the heat pump is the only technology that can provide cooling, this coupling links its size to the choice of the other thermal technologies.

\paragraph{Inverter}
PV, battery and wind turbine are each connected to the AC bus through a dedicated converter: a DC/AC inverter for PV, a bidirectional AC/DC inverter for the battery, and a frequency converter for the wind turbine. Each converter $j\in\{\mathrm{pv},\mathrm{bat},\mathrm{wt}\}$ is modelled as:
\begin{equation}
p^{\text{out}}_{\mathrm{inv},j}(t) = \eta_{\text{inv},j}\, p^{\text{in}}_{\mathrm{inv},j}(t), \qquad 0 \le p^{\text{in}}_{\mathrm{inv},j}(t) \le P_{\text{inv},j}^{\max},
\end{equation}
where $p^{\text{in}}_{\mathrm{inv},j}$ is the power drawn from the source side and $p^{\text{out}}_{\mathrm{inv},j}$ the power delivered to the other side. 
The rating $P_{\text{inv},j}^{\max}$ and the efficiency $\eta_{\text{inv},j}$ are design variables; the converter is installed whenever its technology is installed, and its rating, bounded below by 1~kW, can limit the power that the technology delivers. 
In the case study the converter cost is taken as already included in the specific costs of PV, wind turbine and battery, so the converters carry zero CAPEX and OPEX (Table~\ref{tab:capex}); their rating and efficiency affect the NPC only through the variable operating cost, since conversion losses and clipping reduce the energy delivered and therefore increase grid purchases or reduce exports.

\section{Net Present Cost calculation}\label{app:npc}

This appendix details the terms of the Net Present Cost of Eq.~\mref{eq:npc}.

\textbf{Present Value Factor.} The present value of a future cash flow is calculated using the discount rate $r = 4\%$:
\begin{equation}
    \text{PV}(C, y) = \frac{C}{(1+r)^y}
\end{equation}

For a series of equal annual payments (annuity), the present value factor is:
\begin{equation}
    \text{PV}_{\text{factor}} = \sum_{y=1}^{N} \frac{1}{(1+r)^y} = \frac{(1+r)^N - 1}{r(1+r)^N}
\end{equation}

where $\text{PV}(C,y)$ is the present value of a cash flow $C$ (EUR) occurring at year $y$, $r$ is the annual discount rate (-), $N$ is the project horizon (years), and $\text{PV}_{\text{factor}}$ is the annuity present-value factor (-). In this study, $N = 25$ and $r = 0.04$, yielding $\text{PV}_{\text{factor}} \approx 15.62$.

\textbf{CAPEX with Replacements.} For technologies with lifetime shorter than the project horizon, replacement costs are included:

\begin{equation}
    \text{CAPEX}_{\text{total}} = \sum_{i} \left[ \text{CAPEX}_i + \sum_{y \in \mathcal{R}_i} \frac{\text{CAPEX}_i}{(1+r)^y} \right]
\end{equation}

where $i$ indexes technologies, $\text{CAPEX}_i$ is initial capital cost for technology $i$ (EUR), and $\mathcal{R}_i$ is the set of replacement years for technology $i$. For example, a component with a 20-year lifetime is replaced once, at year 20, in a 25-year project, whereas a component whose lifetime equals or exceeds the horizon is never replaced.

The replacement cost multiplier over the project lifetime is:
\begin{equation}
    f_{\text{repl}}(L, N, r) = 1 + \sum_{m=1}^{\lceil N/L \rceil - 1} \frac{1}{(1+r)^{mL}}
\end{equation}
where $L$ is component lifetime (years), $m$ is the replacement index, and $f_{\text{repl}}$ is a dimensionless multiplier accounting for discounted replacements.

Two accounting conventions should also be made explicit, because they materially affect the ranking of technologies. First, equipment that is still usable at the end of the horizon is credited with its residual value. Without this credit, a component replaced shortly before year $N$ would be charged its full discounted cost although most of its life remains, which systematically penalises short-lived technologies. Second, no decommissioning or end-of-life cost is included. The simulator credits a linear residual value
\begin{equation}
S_i \;=\; \mathrm{CAPEX}_i \cdot \frac{L_i - (N - y_i^{\text{last}})}{L_i} \cdot \frac{1}{(1+r)^{N}},
\label{eq:salvage}
\end{equation}
where $y_i^{\text{last}}$ is the year of the last installation of technology $i$, and subtracts $S_i$ from the NPC. 

The effective CAPEX including replacements is:
\begin{equation}
    \text{CAPEX}_{\text{effective},i} = \text{CAPEX}_i \times f_{\text{repl}}(L_i, N, r)
\end{equation}
where $L_i$ is the lifetime of technology $i$ (years) and $\text{CAPEX}_{\text{effective},i}$ is total discounted investment attributable to technology $i$ over the full horizon (EUR), so that $\text{CAPEX}_{\text{total}}=\sum_i \text{CAPEX}_{\text{effective},i}$.

\textbf{Fixed OPEX.} Annual fixed operational expenditures are converted to present value:
\begin{equation}
    \text{OPEX}_{\text{fixed}}^{\text{PV}} = \sum_{i} \text{OPEX}_{\text{fixed},i} \times \text{PV}_{\text{factor}}
\end{equation}
where $\text{OPEX}_{\text{fixed},i}$ is annual fixed operating cost for technology $i$ (EUR/yr) and $\text{OPEX}_{\text{fixed}}^{\text{PV}}$ is its discounted present value over the project horizon (EUR).

\textbf{Variable OPEX.} Variable operational costs depend on the dispatch solution of the linear program of Section~\mref{sec:dispatch}:
\begin{align}
    \text{OPEX}_{\text{variable}} = \sum_t \Big[ &\, c_{\text{elec}} \cdot p_{\text{grid}}^{\text{import}}(t) - c_{\text{export}} \cdot p_{\text{grid}}^{\text{export}}(t) \nonumber \\
    &+ c_{\text{gas}} \Big( \tfrac{p_{\text{boiler}}(t)}{\eta_{\mathrm{boi}}} + p_{\text{chp}}^{\text{gas}}(t) \Big) + c_{\text{chp}}^{\text{var}} \cdot \,\eta_{\mathrm{chp}}^{\mathrm{el}}\,p_{\text{chp}}^{\text{gas}}(t) \nonumber \\
    &+ \pi \big( s^{\text{el}}(t) + s^{\text{heat}}(t) + s^{\text{cool}}(t) \big) \Big]
\end{align}

where $t$ indexes time steps (hours), $p_{\text{grid}}^{\text{import}}(t)$ and $p_{\text{grid}}^{\text{export}}(t)$ are imported and exported grid electricity (kW), $p_{\text{boiler}}(t)$ is boiler thermal output (kW$_{\text{th}}$), $\eta_{\mathrm{boi}}$ is boiler efficiency (-), $p_{\text{chp}}^{\text{gas}}(t)$ is CHP gas input power (kW$_{\text{fuel}}$), so that $\eta_{\mathrm{chp}}^{\mathrm{el}}p_{\text{chp}}^{\text{gas}}(t)$ is the CHP electrical output (kW$_{\text{el}}$), $c_{\text{elec}}$ is electricity purchase price (EUR/kWh, default 0.30), $c_{\text{export}}$ is the export tariff (EUR/kWh, default 0.08), $c_{\text{gas}}$ is gas price (EUR/kWh, default 0.12), and $c_{\text{chp}}^{\text{var}}$ is the CHP variable maintenance cost (EUR/kWh$_{\text{el}}$, default 0.03). Unserved electricity, heat, and cooling are additionally charged at a penalty of 1000 EUR/kWh, which forces feasibility and exposes under-sizing through the penalty term.

\section{Optimal dispatch formulation}\label{app:dispatch}

\textbf{Decision Variables} (defined for every $t$):
\begin{align}
    p_{\text{grid}}^{\text{import}}(t),\ p_{\text{grid}}^{\text{export}}(t)
        &\quad \text{grid import / export (kW)} \\
    p_{\text{batt}}^{\text{charge}}(t),\ p_{\text{batt}}^{\text{discharge}}(t)
        &\quad \text{battery charge / discharge power (kW)} \\
    p_{\text{boiler}}(t)
        &\quad \text{boiler thermal output (kW$_{\text{th}}$),} \nonumber \\
    p_{\text{chp}}^{\text{gas}}(t)
        &\quad \text{CHP gas input (kW$_{\text{fuel}}$)} \\
    p_{\text{hp}}^{\text{el}}(t),\ p_{\text{hp,c}}^{\text{el}}(t)
        &\quad \text{heat-pump electrical input, heating / cooling (kW)} \\
    \text{soc}(t)
        &\quad \text{battery state of charge (kWh)} \\
    s^{\text{el}}(t),\ s^{\text{heat}}(t),\ s^{\text{cool}}(t)
        &\quad \text{unserved electricity / heat / cooling (kW)} \\
    p_{\text{st}}(t)
        &\quad \text{solar-thermal heat actually used (kW}_{\text{th}}\text{)}
\end{align}

\textbf{Objective Function:}
\begin{align}
    \min \sum_{t=1}^{T} \Big[\,
    & c_{\text{elec}}\, p_{\text{grid}}^{\text{import}}(t)
    - c_{\text{export}}\, p_{\text{grid}}^{\text{export}}(t)
    + c_{\text{gas}} \Big( \frac{p_{\text{boiler}}(t)}{\eta_{\mathrm{boi}}}
    + p_{\text{chp}}^{\text{gas}}(t) \Big) \nonumber \\
    & + c_{\text{chp}}^{\text{var}}\, \eta_{\mathrm{chp}}^{\mathrm{el}}\, p_{\text{chp}}^{\text{gas}}(t)
    + \pi \big( s^{\text{el}}(t) + s^{\text{heat}}(t) + s^{\text{cool}}(t) \big)
    \Big]
\end{align}
where $\pi = 1000$~EUR/kWh is the penalty on unserved demand.

\textbf{Constraints} (for all $t$):
\begin{align}
    & \text{Electric balance:} \nonumber \\
    & \quad p_{\text{pv}}(t) + p_{\text{wind}}(t) + p_{\text{chp}}^{\text{el}}(t)
      + p_{\text{grid}}^{\text{import}}(t)
      + p_{\text{batt}}^{\text{discharge}}(t)
      + s^{\text{el}}(t) \nonumber \\
    & \quad = p_{\text{load}}^{\text{el}}(t)
      + p_{\text{hp}}^{\text{el}}(t)
      + p_{\text{hp,c}}^{\text{el}}(t)
      + p_{\text{grid}}^{\text{export}}(t)
      + p_{\text{batt}}^{\text{charge}}(t) \\[4pt]
    & \text{Heat balance:} \nonumber \\
    & \quad p_{\text{chp}}^{\text{th}}(t) + p_{\text{boiler}}(t)
      + p_{\text{hp}}^{\text{th}}(t) + p_{\text{st}}(t)
      + s^{\text{heat}}(t) \nonumber \\
    & \quad = p_{\text{load}}^{\text{heat}}(t) + p_{\text{load}}^{\text{dhw}}(t) \;\\[4pt]
    & \text{Solar-thermal curtailment:} \nonumber \\
    & \quad 0 \le p_{\text{st}}(t) \le \hat p_{\text{st}}(t)
      = \mathrm{GHI}(t)\,A_{\mathrm{st}}\,\eta_{\text{st}}/1000 \\[4pt]
    & \text{Cool balance:} \nonumber \\
    & \quad p_{\text{hp,c}}^{\text{cool}}(t) + s^{\text{cool}}(t)
      = p_{\text{load}}^{\text{cool}}(t) \\[4pt]
    & \text{CHP coupling:} \nonumber \\
    & \quad 
      p_{\text{chp}}^{\text{el}}(t) = \eta_{\mathrm{chp}}^{\mathrm{el}}\, p_{\text{chp}}^{\text{gas}}(t),\;\;
      p_{\text{chp}}^{\text{th}}(t) = \sigma\, p_{\text{chp}}^{\text{el}}(t),\;\;
      \sigma = \eta_{\mathrm{chp}}^{\mathrm{th}}/\eta_{\mathrm{chp}}^{\mathrm{el}} \\[4pt]
    & \text{HP heating / cooling:} \nonumber \\
    & \quad p_{\text{hp}}^{\text{th}}(t) = \mathrm{COP}\cdot p_{\text{hp}}^{\text{el}}(t),
    \quad
      p_{\text{hp,c}}^{\text{cool}}(t) = \mathrm{EER}\cdot p_{\text{hp,c}}^{\text{el}}(t) \\[4pt]
    & \text{HP shared compressor:} \nonumber \\
    & \quad \frac{p_{\text{hp}}^{\text{el}}(t)}{P_{\text{hp}}^{\max}/\mathrm{COP}} + \frac{p_{\text{hp,c}}^{\text{el}}(t)}{P_{\text{hp}}^{\max}/\mathrm{EER}} \le 1
      \label{eq:hp_timeshare} \\[4pt]
    & \text{Battery SOC:} \nonumber \\
    & \quad \mathrm{soc}(t{+}1) = \mathrm{soc}(t)
      + \eta_{\text{store}}\, p_{\text{batt}}^{\text{charge}}(t)
      - \frac{p_{\text{batt}}^{\text{discharge}}(t)}{\eta_{\text{dispatch}}} \\[4pt]
    & \text{Cyclic SOC:} \nonumber \\
    & \quad \mathrm{soc}(0) = \mathrm{soc}(T) \\[4pt]
    & \text{Capacity limits:} \nonumber \\
    & \quad 0 \le p(t) \le P^{\text{rated}} \quad \forall\,\text{technologies},
      \quad 0 \le \mathrm{soc}(t) \le E_{\text{bat}}^{\max} \\[4pt]
    & \text{Inverter limits:} \nonumber \\
    & \quad 0 \le p^{\text{in}}_{\mathrm{inv},j}(t) \le P_{\mathrm{inv},j}^{\max},
      \quad j\in\{\mathrm{pv},\mathrm{bat},\mathrm{wt}\} \\[4pt]
    & \text{Non-negativity:} \nonumber \\
    & \quad s^{\text{el}}(t),\; s^{\text{heat}}(t),\; s^{\text{cool}}(t)\;\ge 0
\end{align}

The unserved-energy variables $s^{\text{el}}, s^{\text{heat}}, s^{\text{cool}}$, which correspond to those of the problem formulation in Section~\mref{subsec:simulator}, make the balance equalities always satisfiable: any demand that the installed capacity cannot meet is recorded as unserved energy and penalised at $\pi$ in the objective. This guarantees a finite optimum for every design and reveals under-sizing through the penalty cost.

\section{Design variables and cost parameters}\label{app:params}

Table~\ref{tab:design_variables} lists the design variables grouped by technology and Table~\ref{tab:design_bounds} their admissible ranges; Tables~\ref{tab:capex} and~\ref{tab:opex} report the capital costs, lifetimes, replacements and fixed operating costs adopted in the case study.

The decision variables are defined as follows:
\begin{itemize}[noitemsep]
    \item $z_{\mathrm{pv}}, z_{\mathrm{wt}}, z_{\mathrm{chp}}, z_{\mathrm{st}}, z_{\mathrm{boi}}, z_{\mathrm{bat}}, z_{\mathrm{hp}} \in \{0,1\}$: binary activation variables (1 if the technology is installed, 0 otherwise).
    \item $A_{\mathrm{pv}}$ (m$^2$): installed PV area.
    \item $P_{\mathrm{wt}}^{\mathrm{rated}}$ (kW): rated electrical power of the wind turbine.
    \item $P_{\mathrm{chp}}^{\mathrm{el}}$ (kW): nominal electrical power of the CHP unit.
    \item $\eta_{\mathrm{chp}}^{\mathrm{el}}$ (-): electrical efficiency of CHP.
    \item $\eta_{\mathrm{chp}}^{\mathrm{th}}$ (-): thermal efficiency of CHP.
    \item $A_{\mathrm{st}}$ (m$^2$): installed solar-thermal collector area.
    \item $P_{\mathrm{boi}}^{\max}$ (kW$_{\mathrm{th}}$): maximum thermal power of the boiler.
    \item $\eta_{\mathrm{boi}}$ (-): boiler thermal efficiency.
    \item $E_{\mathrm{bat}}^{\max}$ (kWh): maximum battery energy capacity.
    \item $P_{\mathrm{bat}}^{\max}$ (kW): maximum battery charge/discharge power.
    \item $\eta_{\mathrm{bat}}$ (-): round-trip battery efficiency (or equivalent aggregate charge/discharge efficiency used in the model).
    \item $P_{\mathrm{inv},j}^{\max}$ (kW), $\eta_{\mathrm{inv},j}$ (-): nominal power and conversion efficiency of inverter $j\in\{\mathrm{pv},\mathrm{bat},\mathrm{wt}\}$ (three independent inverters).
    \item $P_{\mathrm{hp}}^{\max}$ (kW$_{\mathrm{th}}$): maximum thermal power of the reversible heat pump (shared by heating and cooling).
\end{itemize}

\begin{table}[pos=H]
\centering
\caption{Design variables in $\mathbf{x}$ grouped by technology}
\label{tab:design_variables}
\begin{tabular}{ll}
\toprule
\textbf{Technology} & \textbf{Variables} \\
\midrule
PV solar & $z_{\mathrm{pv}},\ A_{\mathrm{pv}}$ \\
Wind turbine & $z_{\mathrm{wt}},\ P_{\mathrm{wt}}^{\mathrm{rated}}$ \\
CHP & $z_{\mathrm{chp}},\ P_{\mathrm{chp}}^{\mathrm{el}},\ \eta_{\mathrm{chp}}^{\mathrm{el}},\ \eta_{\mathrm{chp}}^{\mathrm{th}}$ \\
Solar thermal & $z_{\mathrm{st}},\ A_{\mathrm{st}}$ \\
Boiler & $z_{\mathrm{boi}},\ P_{\mathrm{boi}}^{\max},\ \eta_{\mathrm{boi}}$ \\
Battery & $z_{\mathrm{bat}},\ E_{\mathrm{bat}}^{\max},\ P_{\mathrm{bat}}^{\max},\ \eta_{\mathrm{bat}}$ \\
Inverter PV & $P_{\mathrm{inv},\mathrm{pv}}^{\max},\ \eta_{\mathrm{inv},\mathrm{pv}}$ \\
Inverter battery & $P_{\mathrm{inv},\mathrm{bat}}^{\max},\ \eta_{\mathrm{inv},\mathrm{bat}}$ \\
Inverter wind & $P_{\mathrm{inv},\mathrm{wt}}^{\max},\ \eta_{\mathrm{inv},\mathrm{wt}}$ \\
Heat pump (reversible) & $z_{\mathrm{hp}},\ P_{\mathrm{hp}}^{\max}$ \\
\bottomrule
\end{tabular}
\end{table}

\begin{table}[pos=H]
\centering
\caption{Admissible ranges and quantization steps of the 18 continuous design
variables, and values of the fixed technology parameters. Source: \texttt{settings.PARAM\_RANGES} and \texttt{settings.py}}
\label{tab:design_bounds}
\small
\begin{tabular}{@{}llrrrc@{}}
\toprule
\textbf{Technology} & \textbf{Variable} & \textbf{Min} & \textbf{Max} & \textbf{Step} & \textbf{Levels} \\
\midrule
PV            & $A_{\mathrm{pv}}$ (m$^2$)                      & 5.0  & 2000.0 & 5.0  & 400 \\
Wind          & $P_{\mathrm{wt}}^{\mathrm{rated}}$ (kW)        & 1.5  & 25.5   & 1.5  & 17 \\
CHP           & $P_{\mathrm{chp}}^{\mathrm{el}}$ (kW)          & 5.0  & 100.0  & 5.0  & 20 \\
CHP           & $\eta_{\mathrm{chp}}^{\mathrm{el}}$ (--)       & 0.24 & 0.32   & 0.01 & 9 \\
CHP           & $\eta_{\mathrm{chp}}^{\mathrm{th}}$ (--)       & 0.50 & 0.58   & 0.01 & 9 \\
Solar thermal & $A_{\mathrm{st}}$ (m$^2$)                      & 5.0  & 500.0  & 5.0  & 100 \\
Boiler        & $P_{\mathrm{boi}}^{\max}$ (kW$_{\mathrm{th}}$) & 5.0  & 1200.0 & 5.0  & 240 \\
Boiler        & $\eta_{\mathrm{boi}}$ (--)                     & 0.80 & 0.99   & 0.01 & 20 \\
Battery       & $E_{\mathrm{bat}}^{\max}$ (kWh)                & 5.0  & 100.0  & 5.0  & 20 \\
Battery       & $P_{\mathrm{bat}}^{\max}$ (kW)                 & 2.5  & 50.0   & 2.5  & 20 \\
Battery       & $\eta_{\mathrm{bat}}$ (--)                     & 0.70 & 0.99   & 0.01 & 30 \\
Inverter PV   & $P_{\mathrm{inv,pv}}^{\max}$ (kW)              & 1.0  & 400.0  & 1.0  & 400 \\
Inverter PV   & $\eta_{\mathrm{inv,pv}}$ (--)                  & 0.95 & 0.98   & 0.01 & 4 \\
Inverter bat. & $P_{\mathrm{inv,bat}}^{\max}$ (kW)             & 1.0  & 50.0   & 1.0  & 50 \\
Inverter bat. & $\eta_{\mathrm{inv,bat}}$ (--)                 & 0.95 & 0.98   & 0.01 & 4 \\
Inverter wind & $P_{\mathrm{inv,wt}}^{\max}$ (kW)              & 1.0  & 25.0   & 1.0  & 25 \\
Inverter wind & $\eta_{\mathrm{inv,wt}}$ (--)                  & 0.95 & 0.98   & 0.01 & 4 \\
Heat pump     & $P_{\mathrm{hp}}^{\max}$ (kW$_{\mathrm{th}}$)  & 50.0 & 1100.0 & 50.0 & 22 \\
\midrule
\multicolumn{6}{@{}l@{}}{\emph{Fixed technology parameters}:\;
$\eta_{\mathrm{pv}}=0.18$;\;
$\eta_{\mathrm{st}}=0.70$;} \\
\multicolumn{6}{@{}l@{}}{\phantom{\emph{Fixed technology parameters}}\;\;\;
$v_{\mathrm{in}}=3$~m/s, $v_{\mathrm{rated}}=12$~m/s, $v_{\mathrm{out}}=25$~m/s;} \\
\multicolumn{6}{@{}l@{}}{\phantom{\emph{Fixed technology parameters}}\;\;\;
COP $=3.5$;\; EER $=3.0$;\; $\pi=1000$~EUR/kWh;} \\
\multicolumn{6}{@{}l@{}}{\phantom{\emph{Fixed technology parameters}}\;\;\;
$r=0.04$;\; $N=25$~yr.} \\
\bottomrule
\end{tabular}
\end{table}

\begin{table}[pos=H]
\centering
\caption{CAPEX and Lifetime Parameters}
\label{tab:capex}
\setlength{\tabcolsep}{4pt}
\begin{tabular}{lcccc}
\toprule
Technology & Unit & CAPEX (EUR/unit) & Lifetime (yr) & Replacements \\
\midrule
PV & m$^2$ & 121.3 & 40 & 0 \\
Micro Wind & kW & 4,040 & 20 & 1 \\
CHP & kW$_{\text{el}}$ & 1,010.2 & 25 & 0 \\
Solar Thermal & m$^2$ & 443 & 30 & 0 \\
Boiler & kW$_{\text{th}}$ & 158.2 & 25 & 0 \\
Battery & kWh & 180 & 20 & 1 \\
Inverters & kW & 0 (incl.) & 40 & 0 \\
Heat Pump & kW$_{\text{th}}$ & 918.7 & 20 & 1 \\
\bottomrule
\end{tabular}
\end{table}

\begin{table}[pos=H]
\centering
\caption{Fixed OPEX Parameters}
\label{tab:opex}
\begin{tabular}{lcc}
\toprule
Technology & Unit & OPEX (EUR/unit/yr) \\
\midrule
PV & m$^2$ & 0.44 \\
Micro Wind & kW & 101 \\
CHP & kW$_{\text{el}}$ & 10.37 \\
Solar Thermal & m$^2$ & 2.36 \\
Boiler & (flat) & 692.7 \\
Battery & kWh & 3.60 \\
Inverter & kWac & - \\
Heat Pump & kW$_{\text{th}}$ & 18.13 \\
\bottomrule
\end{tabular}
\end{table}

\section{Consumption and weather data}\label{app:data}

The profiles used in the experiments are not synthetic but are constructed from measured hourly electricity consumption and hourly weather observations for the year 2024 (8,784 hourly samples, including 29 February).

The following hourly series are used by the simulator:
\begin{itemize}[noitemsep]
    \item Global Horizontal Irradiance (W/m$^2$), representing solar resource availability
    \item Wind speed at 10m height (m/s), driving wind turbine output
    \item Electrical demand (kW), pure electrical residual of the measured load
    \item Space heating demand (kW$_{\text{th}}$), temperature-dependent thermal load
    \item Space cooling demand (kW$_{\text{th}}$), temperature-dependent summer load
    \item Domestic hot water demand (kW$_{\text{th}}$), occupancy-driven baseload
    \item measured total electrical load (kW), kept for reference
\end{itemize}

The weather series (solar irradiance, wind speed, air temperature) are obtained from the Open-Meteo reanalysis for the coordinates of the site (40.88$^{\circ}$N, 14.34$^{\circ}$E; hourly shortwave radiation, 2~m air temperature and 10~m wind speed, year 2024).

Table \ref{tab:profile_stats} summarizes the key statistics of the load and resource profiles.

\begin{table}[pos=H]
\centering
\caption{Profile Statistics (Annual)}
\label{tab:profile_stats}
\begin{tabular}{lcccc}
\toprule
Variable & Mean & Peak & Minimum & Total (MWh) \\
\midrule
GHI (W/m$^2$) & 189 & 937 & 0 & -- \\
Wind speed (m/s) & 2.2 & 10.4 & 0 & -- \\
Air temperature (°C) & 17.6 & 36.9 & 0.6 & -- \\
Elec. demand (kW) & 80.0 & 255.4 & 34.8 & 703.1 \\
Heat demand (kW$_{\text{th}}$) & 52.3 & 630.6 & 0 & 459.0 \\
Cool demand (kW$_{\text{th}}$) & 31.5 & 663.6 & 0 & 276.8 \\
DHW demand (kW$_{\text{th}}$) & 24.4 & 146.6 & 0 & 214.5 \\
Measured total load (kW) & 112.5 & 336.0 & 34.8 & 987.8 \\
\bottomrule
\end{tabular}
\end{table}

\begin{figure}[pos=H]
\centering
\includegraphics[width=0.8\textwidth]{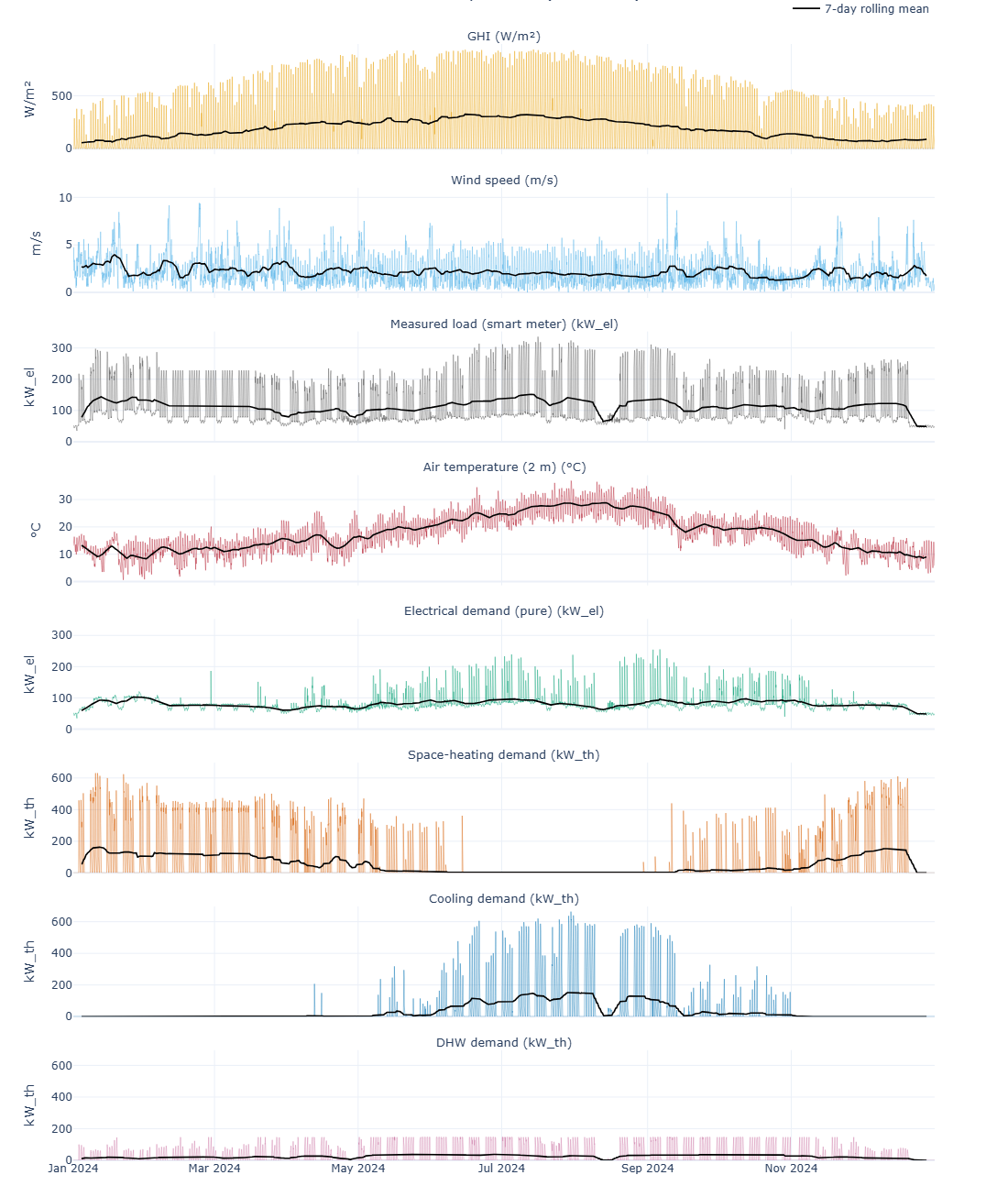}
\caption{Annual demand and resource profiles used in the case study.}
\label{fig:profiles}
\end{figure}

Figure~\ref{fig:profiles} provides a visual overview of the resulting annual profiles, showing the temporal structure of electricity demand, thermal demand components, and weather-dependent renewable resources used by the simulator. 

The measured series is an \emph{electrical} load that includes the electricity used for space heating, space cooling and domestic hot water, in addition to the pure electrical uses. Since these end uses are not metered separately, the split is estimated. Each thermal-related share is given an hourly shape (heating and cooling degree-hours, with base temperatures of 20 and 22~$^{\circ}$C, during weekday office hours; a flat profile over weekday occupancy for DHW) and scaled to an assumed fraction of the annual load. 
In every hour the three shares are then capped so as to leave at least a daily baseline of pure electrical use, estimated from the night-time load. 
Since the two base temperatures do not overlap, space heating and cooling never occur in the same hour, whereas DHW can coincide with cooling.

\section{Qualitative analysis of the reasoning traces}\label{app:qualitative}

In the following we report how the sessions comply with the rules of Section~\mref{subsubsec:instructions}, and the reasoning error that explains the two Sonnet~5.5 runs that never come within 1\% of the best design.

No run reads or modifies the simulator, the cost parameters or the hooks, or writes the experiment counter or the state files; every simulation is launched through the hooks, and no run records an infeasible design as accepted. The rule of changing one to three logically related parameters per experiment is respected by all runs, if switching a technology on or off is counted as a single change: a technology that is switched on must also be given a size, and for PV and the battery an inverter, to be tested at all. Counted parameter by parameter, 32 of the 500 Opus experiments, 31 of the 500 Sonnet experiments and 16 of the 485 Haiku experiments change four or more parameters; all of them switch on a technology (CHP, battery, boiler, wind turbine or PV) together with its size, efficiency or inverter, in ten cases a boiler together with a change of the heat-pump size. Seven simulations (three by Opus, three by Sonnet and one by Haiku) are run on an unchanged configuration and are recognised and not recorded by the post-execution hook. One Opus run repeats a rejected configuration, against the rules, while one Opus run and two Haiku runs repeat four accepted configurations in total, which wastes budget but is not forbidden. In the ablation without \texttt{insights.md} (Section~\mref{subsec:ablation_insights}), no session reads the file, and a single experiment, in a Haiku~5.5 run, violates the rule, changing the battery capacity, power and inverter together with the heat-pump size.

In one of the two trapped Sonnet runs, the agent reduces the heat pump from 800 to 750~kW at experiment~13, in a design without other heat sources; the 750~kW heat pump leaves 2.3~kWh of cooling unserved, and the agent concludes that ``800~kW is the minimum heat pump size''; in the other run the same shortfall appears at experiment~6, again with no other heat source, and the agent states that ``the heat pump at 750~kW is right at the cooling peak limit'', moves to 800~kW and never tests 700~kW. 
Both attribute the shortfall to the cooling peak (664~kW\textsubscript{th}), whereas it is caused by the coincidence of cooling and hot-water demand (Eq.~\eqref{eq:hp_timeshare} in Appendix~\ref{app:dispatch}), and neither retests the heat pump after CHP and solar thermal have been added; both end by declaring the search ``Fully converged'' or ``complete''. 
The Opus runs that face the same situation retest the heat pump once the other heat sources are in place, and reach 700~kW.
Haiku~5.5 shows errors of the same kind, such as discarding the CHP after testing it only at its default efficiencies, crediting the battery for making a 700~kW heat pump feasible, and frequent claims of convergence while a better design exists.

\ifx\inmain\undefined\def\next{\end{document}}\else\let\next\relax\fi
\next

\end{document}